\documentclass[10pt,twocolumn,letterpaper]{article}
\PassOptionsToPackage{table}{xcolor}
\usepackage[pagenumbers]{lacepreprint}
\usepackage[utf8]{inputenc}
\usepackage[T1]{fontenc}
\usepackage{amsthm}
\usepackage{array,tabularx,multirow,threeparttable,adjustbox}
\usepackage{algorithm,algpseudocode}
\usepackage{float}
\usepackage{microtype}
\definecolor{linkblue}{rgb}{0.21,0.49,0.74}
\usepackage[breaklinks,colorlinks,allcolors=linkblue,pdfusetitle]{hyperref}
\graphicspath{{../}}
\providecommand{\cmark}{$\bullet$}
\providecommand{\omark}{$\circ$}
\providecommand{\xmark}{$\times$}

\title{Latent Commonality Expectation-Maximisation for Box-supervised Tree Crown Instance Segmentation}
\author{Thomas Pitts\thanks{ORCID \href{https://orcid.org/0009-0000-8884-2381}{0009-0000-8884-2381}.}\qquad Kunqi Li\qquad Bin Liang\thanks{Corresponding author.}\\
Data Science Institute, University of Technology Sydney, Sydney, NSW 2007, Australia\\
{\tt\small bin.liang@uts.edu.au}}
\hypersetup{pdfauthor={Thomas Pitts, Kunqi Li, Bin Liang}}

\begin{document}
\maketitle
\begin{abstract}
Individual tree crown segmentation from aerial imagery yields tree-level inventories of woody vegetation, underpinning carbon stock estimation, biodiversity assessment and restoration monitoring at landscape scale. However, existing models are predominantly trained on dense-canopy forest imagery and degrade in savannas and drylands, where tree crowns are sparse, of variable appearance, and underrepresented in annotated benchmarks. We introduce LACE (LAtent Commonality Expectation-maximisation), a box-supervised instance segmentation model for very-high-resolution ($0.1$ m/px) aerial imagery. Its novel decoder-free module learns crown appearance from the commonality of embeddings across training boxes, contrasted against their surroundings. This approach distils a latent ``treeness'' from the feature space of a frozen DINOv3-web Vision Transformer (ViT-L/16) encoder to derive masks from the boxes of a DINO-DETR detector. On the multi-biome OAM-TCD holdout test set, LACE trained on $900$ box-annotated images outperforms the OAM-TCD authors' Mask R-CNN trained on ${\sim}4.2$k polygon-annotated images (mask average precision, AP$_{50}$, $0.714$ vs $0.648$) with an estimated $14\times$ fewer annotator-hours, as well as the strongest evaluated box-supervised baseline, BoxTeacher ($0.664$). It also leads every model evaluated on $236$ unseen sparse-canopy tiles ($0.743$ vs BoxTeacher's $0.721$) and, given identical boxes, its module outperforms the mask-pretrained Segment Anything Model~3 ($0.714$ vs $0.701$). LACE demonstrates that accurate tree crown segmentation no longer requires polygon masks, opening crown mapping to sparse-canopy landscapes where annotated datasets remain scarce.
\end{abstract}

\section{Introduction}

Individual tree crowns capture the horizontal structure of woody vegetation, and their accurate measurement can serve as a proxy for woody biomass via allometries~\cite{hiernaux2023,tucker2023}, thereby enabling monitoring of carbon stock within areas containing trees. Measurement techniques in this domain have traditionally focused on dense canopy areas and forests, but there is an increasing body of literature that highlights the limitations of this approach as distribution, density, cover, and carbon content of trees in sparse-canopy biomes (listed in Table~\ref{tab:sparse-biomes}) are not well understood at sub-continental to continental scales~\cite{tucker2023,selvabox,fagan2020}. Recent studies highlight that the global contribution of sparse-canopy biomes as a carbon sink is underestimated due to the underestimation of tree mass in sparse-canopy areas~\cite{bastin2017,tucker2023,brandt2020,skole2021}, even while semi-arid ecosystems account for roughly half of the trend in the global land carbon sink and $39$--$47\%$ of its interannual variability~\cite{ahlstrom2015}. At the same time, a quarter of all trees globally are located in grassland, savanna, desert and tundra biomes~\cite{crowther2015}. Drylands cover ${\sim}41.5\%$ of the Earth's land surface~\cite{sorensen2007,bastin2017} and are expanding further: they are projected to cover $50\%$ or more by 2100~\cite{huang2016}.

Machine learning and deep learning applied to satellite and unmanned aerial vehicle (UAV) data have shifted the field from aggregate cover fractions toward inventories of individual trees~\cite{soomro2026}: Brandt et al.~\cite{brandt2020} mapped over $1.8$ billion crowns across $1.3$ million $\mathrm{km}^2$ of the Sahara and Sahel from $0.5$~m imagery. Tucker et al.~\cite{tucker2023} extended this to carbon estimation over $9.9$ billion individual trees. Mugabowindekwe et al.~\cite{mugabowindekwe2023} produced nation-wide tree-level carbon estimates for Rwanda using $0.25$~m/px imagery, and found that $48.6\%$ of aboveground tree carbon stock was located outside forests. They also found that ``even a very detailed manual \emph{forest} delineation\ldots missed $38.4\%$ of the isolated trees in Rwanda, which account for $25.5\%$ of the national aboveground carbon stocks'' (emphasis added). These studies establish that individual crown detection in open, scattered-tree environments is tractable, but, crucially, typically depend on a) (often proprietary) very-high-resolution (VHR, sub-metre) imagery~\cite{skole2021} and b) labelled data~\cite{ghasemi2022}, motivating more transferable and label-efficient methods.

\begin{figure}[t]
\centering
\includegraphics[width=\linewidth]{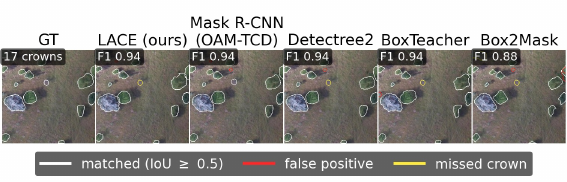}
\caption{Open savanna holdout tile (OAM-TCD $397$, $512$~px crop). Its LACE$-$Mask R-CNN (OAM-TCD) AP$_{50}$ margin ($+0.020$) lies below the open-ground median ($+0.048$). Each model is shown at its own best-F1 threshold.\label{fig:qual}}
\end{figure}

Table~\ref{tab:gap} (Appendix~\ref{app:gap}) summarises representative deep-learning methods for tree crown instance segmentation (TCIS) on red--green--blue (RGB) imagery. We address three gaps. First, existing approaches focus on dense-canopy forests and miss many trees outside them~\cite{mugabowindekwe2023}. For example, the authors of Detectree2 deliberately exclude training and test tiles with less than $40\%$ crown cover~\cite{ball2023}, and in the seminal DeepForest paper the authors highlight the worst-performing site, Onaqui, Utah, as a ``desert scrub site with a different vegetation structure from any of the training data''~\cite{weinstein2020}. Our setting is instead savanna, scrub and other open landscapes, where crowns are sparse, small and set against bare or grassy ground (Figure~\ref{fig:qual}). Second, existing models typically rely on manually annotated polygons for training~\cite{hao2021,braga2020,ocer2020,ball2023,chadwick2020,zhao2023}. Instead, we use boxes only, which have the benefit of being much faster and therefore cheaper to produce~\cite{jain2013,box2mask}. Third, like-for-like comparison across this literature remains difficult, as papers variously report average precision (AP$_{50}$, AP$_{50:95}$), instance- or semantic-F1 at differing intersection over union (IoU) thresholds against boxes and/or polygons. The most widely used matching criterion in TCIS is an IoU $\geq0.5$ between predicted and ground truth (GT) crowns~\cite{ball2023,weinstein2020,hao2021,plesoianu2020}, and we adopt it for our headline metrics, mask AP$_{50}$ and F1.

We see two contributing factors to the exclusion of sparse-canopy biomes. The first is spatial resolution: individual tree crown (ITC) delineation requires VHR imagery, ideally ${\leq}0.1$~m/px as from UAV surveys~\cite{oamtcd}, beyond most publicly available satellite RGB at $5$--$10$~m/px (e.g.\ Sentinel-2~\cite{drusch2012}). To illustrate, downsampling the OpenAerialMap Tree Crown Delineation (OAM-TCD) holdout set used in this work ($439$ images, $25{,}705$ crowns) from its native $0.1$~m/px to a $1$~m Ground Sample Distance (GSD) leaves $25.0\%$ of crowns enclosing no complete pixel. At $1$~m/px, $1$ in $4$ crowns in this representative dataset would therefore be effectively undetectable. 

The second is that conventional global tree cover maps such as Hansen et al.~\cite{hansen2013} systematically underestimate tree cover in drylands~\cite{fagan2020} and are markedly less accurate outside closed-canopy forests~\cite{brandtstolle2021}, while biomass and carbon products used for drylands~\cite{baccini2012,avitabile2016,bouvet2018} typically apply a single method to forest and dryland vegetation alike~\cite{tucker2023}. We speculate that this has led to a self-reinforcing cycle: under-counting makes woody biomass in sparse-canopy biomes appear too scarce to justify annotation effort, which in turn hinders tree crown detection and segmentation in exactly the biomes where cover is most uncertain~\cite{bastin2017}. Tucker et al.~\cite{tucker2023} found that ``areas with scattered trees are often represented by zero values'', since ``previous models are rarely developed, trained and validated with plots of very sparse tree cover''. Since TCIS models remain overwhelmingly supervised (Table~\ref{tab:gap}), datasets precede models. This is beginning to change: SelvaBox~\cite{selvabox} and SelvaMask~\cite{selvamask} have broadened geographic coverage of dense-canopy annotations, and OAM-TCD~\cite{oamtcd} includes a variety of biomes, including urban, tundra and savanna.

Apart from limitations within sparse-canopy coverage, TCIS also remains understudied due to the lack of individual tree crown mask annotations in datasets~\cite{teng2025}. Since crown polygons are slower and typically more expensive to annotate than boxes, box-supervised instance segmentation~\cite{boxinst,box2mask,boxteacher} offers a practical route to instance-level delineation without polygon labels. The recent rise of self-supervised backbones such as DINOv2~\cite{dinov2} and DINOv3~\cite{dinov3} has demonstrated that models pretrained on very large sets of diverse web imagery are capable of matching or even exceeding Earth-observation-specific foundation models (e.g. DINOv3-sat) on high-resolution RGB remote sensing (RS) tasks~\cite{dinov3}. 

\begin{figure*}[t]
\centering
\includegraphics[width=\linewidth]{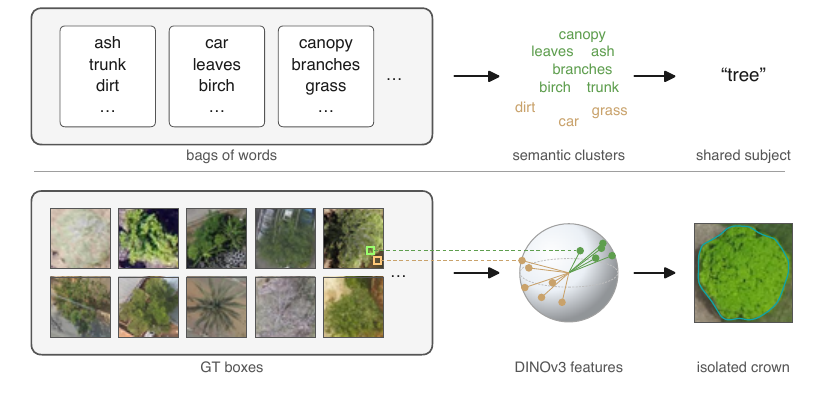}
\caption{Intuition behind LACE. Given highly compressed verbal representations (words), the shared subject can be inferred even though some entries (``dirt'', ``car'') are unrelated (top). Likewise, given only box crops of diverse tree crowns, LACE infers the correct target from groups of features (bottom; see Figure~\ref{fig:masker-fit}). Crops are OAM-TCD training boxes, including man-made objects such as roads and cars.\label{fig:concept}}
\end{figure*}

Strong self-supervised features present the possibility of leveraging the encoding of annotated boxes in representation space to learn by virtue of repeated similarities across annotation boxes. This is conceptually similar to inferring the subject of a list of verbal representations (Figure~\ref{fig:concept}): ``ash'', ``trunk'', ``dirt'', ``car'', ``leaves'', ``birch'', ``canopy'', ``branches'' etc. Challenges remain, not least filtering out unwanted elements such as ``dirt'' and ``car''. Producing a reliable model for the specific task of TCIS is the main aim of this work. We therefore ask whether the vector space signature of the tree-ness within DINO latent feature vectors, based only on the commonality between GT boxes, can be leveraged to isolate the trees from the background. Furthermore, this approach needs to work without the model ever being told which elements of the bounding boxes it should be targeting: the method is tree-agnostic in principle, and only becomes a TCIS model through training exposure. 

Our contributions are threefold:
\begin{itemize}
\item A novel, decoder-free expectation-\hspace{0pt}maximisation (EM) box-to-mask module that isolates crowns in frozen DINOv3-web Vision Transformer (ViT-L/16)~\cite{dinov3} features from the latent commonality of the GT boxes alone. Given identical detector boxes, the module outperforms the mask-pretrained foundation model Segment Anything Model~3 (SAM~3) at AP$_{50}$, and BoxTeacher's own mask head on BoxTeacher's boxes (Section~\ref{sec:samebox}).
\item A finding that a box-supervised model can outperform the strongest released mask-supervised TCIS models: LAtent Commonality Expectation-maximisation (LACE), trained on $900$ box-labelled images, beats the OAM-TCD authors' Mask R-CNN, trained on $4{,}169$ polygon-labelled images, at mask AP$_{50}$ ($0.714$ vs $0.648$), mask AP$_{50:95}$ ($0.342$ vs $0.333$) and instance F1 ($0.700$ vs $0.688$), with $14\times$ fewer annotator-hours.
\item A benchmark comparison of the key box- and mask-supervised state-of-the-art (SOTA) models in the TCIS space, every one re-run and re-scored by us under a single protocol (Appendix~\ref{app:protocol}), and costed in annotator-hours: LACE needs $14\times$ fewer annotator-hours than the OAM-TCD authors' Mask R-CNN while scoring $0.066$ higher at AP$_{50}$ (Figure~\ref{fig:cost}).
\end{itemize}

The remainder of this paper is organized as follows. Section~\ref{sec:related} reviews related work. Section~\ref{sec:LACE} describes LACE, centred on the box-to-mask module (Section~\ref{sec:masker}) and its posterior-derived confidence (Section~\ref{sec:scorer}). Section~\ref{sec:results} benchmarks LACE against box- and mask-supervised baselines on OAM-TCD, costs each in annotator-hours (Section~\ref{sec:cost}), compares box-to-mask modules on identical boxes (Section~\ref{sec:samebox}), tests zero-shot transfer (Section~\ref{sec:phenobench}) and ablates its components (Section~\ref{sec:ablation}). Section~\ref{sec:discussion} discusses limitations and implications, and Section~\ref{sec:conclusions} concludes.

\section{Related Work}
\label{sec:related}

In recent years there has been a focus in the RS community on RGB datasets which can directly leverage Computer Vision deep learning models~\cite{zhu2017} especially through the use of Convolutional Neural Networks (CNNs)~\cite{lecun1998,alexnet} and Vision Transformers (ViT)~\cite{vit}. Mask R-CNN~\cite{maskrcnn}, the instance segmentation member of the region-based CNN (R-CNN) family of models~\cite{rcnn}, extends the Faster R-CNN~\cite{fasterrcnn} detector with a parallel per-instance mask head and is a common architecture for TCIS, including two of our baselines, Detectree2~\cite{ball2023} and the OAM-TCD authors' released Mask R-CNN~\cite{oamtcd} (hereafter Mask R-CNN (OAM-TCD)), both fine-tuned from Detectron2~\cite{detectron2} checkpoints trained on Common Objects in Context (COCO)~\cite{coco} from ImageNet-1k-pretrained~\cite{imagenet} backbones. Alternative representations such as star-convex polygons~\cite{tong2025}, gradient flows~\cite{fgtreeseg} and edge maps~\cite{ghasemi2022} have also been explored to improve crown delineation.

\subsection{Pretrained Foundation Models}

Meta's Segment Anything Model (SAM)~\cite{sam}, now in its third iteration~\cite{sam3}, is a promptable class-agnostic segmentation foundation model. In remote sensing, SAM has been benchmarked zero-shot on satellite imagery~\cite{ren2024space}, and RSPrompter~\cite{rsprompter} overcomes SAM's reliance on prompts such as points/boxes/masks by using a trained prompter to generate prompt embeddings for aerial imagery directly. SAM has further been applied to TCIS either as a box-to-mask module prompted by a separate detector or centre predictor~\cite{fmsam,chen2025zeroshot,treepseco}, or adapted to the domain by fine-tuning, side-tuning or elevation inputs~\cite{teng2025,treesam,ginkgo}.

DINOv3~\cite{dinov3}, the latest of Meta's DINO family~\cite{dino,dinov2}, is a SOTA encoder trained with fully Self-Supervised Learning (SSL) on $1.7$ billion images. Highly context-agnostic, DINOv3-web is an example of a frozen SSL backbone that can serve as a ``universal visual encoder'', which we here apply to aerial RGB imagery. A DINOv3 Satellite model, trained on a satellite image dataset of $493$ million images, has also been released~\cite{dinov3}, but the authors found that the DINOv3-web model generally outperformed it on a variety of large-scale, high-resolution remote sensing benchmarks for segmentation and detection, commenting that these results show that ``general-purpose SSL can match or exceed satellite-specific approaches for tasks that depend on precise object boundaries (segmentation or object detection)''.

\subsection{Box-supervised Instance Segmentation}

Existing approaches fall broadly into two paradigms. The first is loss design with hand-crafted objectives computed from the box and low-level image cues. BoxInst~\cite{boxinst} established the modern approach with a projection loss requiring the projections of the predicted mask to match the box edges and a pairwise affinity loss encouraging neighbouring pixels of similar colour to share a label, reaching $33.2$ mask AP on COCO~\cite{coco} against $39.1$ for its fully mask-supervised counterpart. BoxSnake~\cite{boxsnake} instead supervises polygon vertices directly, and Box2Mask~\cite{box2mask} evolves a level-set function inside each box with an energy over both the image and its deep features, reaching $42.4$ mask AP on COCO test-dev with a Swin-L~\cite{swin} backbone, the highest published box-supervised COCO result we are aware of. The second paradigm is to utilise mask predictions for self-training. An early example is ``Simple Does It''~\cite{khoreva2017}, which generated pseudo-masks inside boxes with GrabCut~\cite{rother2004} and MCG~\cite{pont2017mcg} and recursively trained a segmentation network on them. In BoxTeacher~\cite{boxteacher}, a teacher predicts confidence-weighted pseudo-masks that supervise the student alongside BoxInst's losses, and is updated as a moving average of the student. BoxSeg~\cite{boxseg} extends this pseudo-labelling with quality-aware filtering of the pseudo-masks, reaching $40.9$ COCO mask AP with a ResNet-101 backbone, but has no public code. DiscoBox~\cite{discobox} instead uses a structured energy teacher whose pairwise and cross-image potentials refine masks within boxes. We use one representative of each paradigm, Box2Mask and BoxTeacher, as baselines (Table~\ref{tab:oamtcd439}); to the best of our knowledge, trained box-supervised instance segmentation architectures, as distinct from box-prompted foundation models such as SAM, have not been previously explored for TCIS.

LACE represents a third way, instead building on self-supervised embeddings to derive statistical relationships of in-box characteristics. Unlike EM-Adapt~\cite{emadapt}, which amortises the E-step into a gradient-trained CNN with implicit cross-box sharing, we fit an explicit closed-form commonality mixture over frozen features. This thematically follows DiscoBox~\cite{discobox} and DDT~\cite{ddt} in exploiting similarities between embeddings within GT boxes, but aligns more closely with STEGO~\cite{stego}, eschewing pixel-space calculations in favour of mask inference in latent space. Unsupervised object discovery on DINO features (LOST~\cite{lost}, TokenCut~\cite{tokencut}, deep spectral methods~\cite{deepspectral}, CutLER~\cite{cutler}) also separates objects in latent space, but from the saliency of a single image; LACE instead learns what recurs across many annotated boxes.

\section{Materials \& Methods}
\label{sec:LACE}

Our method produces instance masks for individual tree crowns using bounding-box supervision only. It has three stages: an object detector trained on bounding boxes (Section~\ref{sec:detector}); a dense feature grid obtained by interlacing sixteen offset passes of a frozen DINOv3-web ViT (Section~\ref{sec:features}); and a novel box-to-mask module that converts each predicted box into a mask (Section~\ref{sec:masker}). Predictions are then ranked by a mask-aware confidence that combines the detector score with the mask module's own posterior, refined by a small learned scorer $\hat s$ (Section~\ref{sec:scorer}). Further details are given in Appendix~\ref{app:method}.

\begin{figure*}[t]
\centering
\vspace{6pt}
\includegraphics[width=\textwidth]{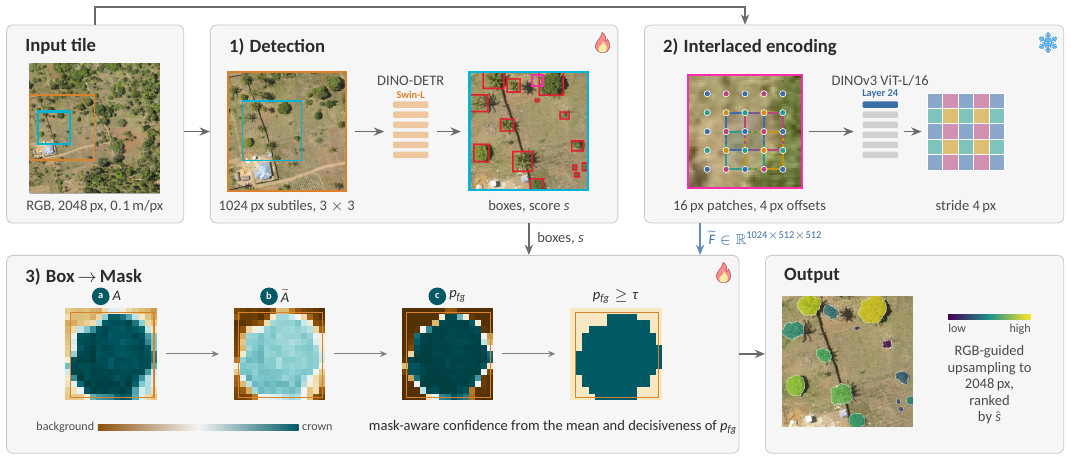}
\vspace{2pt}
\caption{LACE inference pipeline. Boxes from a box-trained detector (1) are masked by a single E-step of the box-to-mask module (3) on interlaced frozen DINOv3 features at stride 4 (2), then upsampled with an RGB-guided filter and ranked by mask-aware confidence $\hat s$.\label{fig:architecture}}
\end{figure*}

\subsection{Box Detector}
\label{sec:detector}

Detection uses the DEtection TRansformer (DETR)-style detector DINO~\cite{dinodetr}, which we write DINO-DETR to distinguish it from the self-supervised DINO encoders, with a Swin-L~\cite{swin} backbone. We initialise it from generic COCO-trained weights and train it with a single tree class on the bounding boxes of the $792$ training images only. Where OAM-TCD marks a region as canopy, a predicted box with $\geq 50\%$ of its area inside the canopy is ignored, mirroring the crowd rule used in evaluation. Each $2048$~px tile is processed as a $3 \times 3$ grid of overlapping $1024$~px subtiles. Boxes scoring below $0.05$ are dropped. The detector score $s$ of each box is DINO-DETR's class probability. Optimisation and stitching details are given in Table~\ref{tab:config}.

\subsection{Dense Features by Phase Interlacing}
\label{sec:features}

A ViT with patch size $p$ emits one embedding per $p \times p$ patch, giving a feature grid $p$ times coarser than the image. We evaluate the frozen encoder sixteen times per tile, shifting the input by $(\delta_y, \delta_x) \in \{0, p/4, p/2, 3p/4\}^2$, and interlace the sixteen grids. With $p = 16$ this yields a $4$~px lattice of $16$~px embeddings, each centred on a separate $4$~px patch. Writing $F^{(\delta_y,\delta_x)}$ for the grid from one offset, the assembled feature tensor $\widetilde F$ is

\begin{equation}
\begin{split}
\widetilde F\!\left[:,\; 4i + \tfrac{\delta_y}{4},\; 4j + \tfrac{\delta_x}{4}\right] &= F^{(\delta_y,\delta_x)}[:,\, i,\, j], \\
i, j &\in \{0, \dots, 127\}
\end{split}
\label{eq:interlace}
\end{equation}

so that cell $(Y, X)$ of $\widetilde F$ carries the embedding of the patch centred at pixel $(4X + 8,\, 4Y + 8)$. Each cell is therefore the encoder output at its own location, rather than any kind of interpolation of its neighbours. The box-to-mask module is fitted on the $8$~px lattice of the four offsets $\{0, p/2\}^2$. All reported results use sixteen offsets at inference, but four offsets, interpolated to the $4$~px lattice, match them at AP$_{50}$ with a quarter of the encoder passes (Section~\ref{sec:ablation}), and we recommend four as the default.

Encoding uses DINOv3 ViT-L/16 with weights frozen throughout. On OAM-TCD we take final-layer (layer~$24$) activations ($1024$ dimensions), giving $\widetilde F \in \mathbb{R}^{1024 \times 512 \times 512}$ for a $2048$~px tile.

\subsection{Box-to-mask Module}
\label{sec:masker}

Given a detection box, the task remains to decide which of its cells belong to the crown. The challenge is that a valid box will almost always contain pixels of the ground or surrounding objects as well, because crowns are not rectangular. We treat this as foreground/background discrimination in a whitened feature space, with the model fitted once on training tiles (using only boxes; Algorithm~\ref{alg:fit}) and then held fixed.

The module has four key components: a whitened feature space; a frozen background mixture; a set of foreground prototypes; and a spatial prior over relative position within the box. The last two are learned by an alternating EM-based fit, while the first two stay fixed.

LACE's core idea is to treat the rich DINOv3-web embeddings as sufficiently descriptive representations of each patch to build a high-performing tree-crown instance segmentation model. We begin with $a$, a given cell's raw unmodified embedding, and whiten it to $d = 128$ dimensions by principal component analysis (PCA). A final $L_2$ normalisation then places the result on the unit sphere, giving the unit vector $z$.

\begin{equation}
\tilde z \;=\; \frac{(a - \mu)\,W}{\lambda}, \qquad z \;=\; \frac{\tilde z}{\lVert \tilde z \rVert}
\label{eq:whiten}
\end{equation}

Here $a \in \mathbb{R}^{1024}$ is one cell of $\widetilde F$; $\mu \in \mathbb{R}^{1024}$ is the cell mean, and $W \in \mathbb{R}^{1024 \times d}$ is the whitening matrix, whose columns are the retained principal directions. $\lambda \in \mathbb{R}^{d}$ is the vector of per-component standard deviations, which are all estimated once on training cells and then frozen. The unit norm makes all subsequent comparisons simple cosine similarities and allows both foreground and background to be modelled by von Mises--Fisher components~\cite{banerjee2005vmf} with a shared concentration $\kappa$.

Cells outside every box, excluding OAM-TCD's canopy class, are background. A background mixture of $K_{bg}$ components $\{(w_j, G_j)\}_{j=1}^{K_{bg}}$ is fitted to them by spherical $k$-means and \emph{frozen}, giving the background log-likelihood $\ell_{bg}$ as follows:

\begin{equation}
\ell_{bg}(z) \;=\; \log \textstyle\sum_j w_j \exp\!\left(\kappa\, z^{\top} G_j\right)
\label{eq:bgll}
\end{equation}

The log-sum-exp acts as a smooth maximum over the components, so $\ell_{bg}(z)$ is a single scalar per cell reading as how well that cell matches its best-matching background component.

Foreground takes the same von Mises--Fisher form, but its weights are supplied per cell by the spatial prior and re-estimated at every M-step, as opposed to global and frozen. Foreground is represented by $K_{fg}$ unit prototypes $\{C_k\}_{k=1}^{K_{fg}}$, initialised from in-box cells atypical of the background. This heuristic initialises the fit. Without this restriction, the initialisation is seeded from in-box background elements such as soil and roads, which are numerous and not the intended crown foreground. In practice the contrastive M-step (Equation~\eqref{eq:contrastive}) draws the prototypes onto a single shared direction, the same at every EM seed (Table~\ref{tab:collapse}): the mixture is how LACE finds one `treeness' direction without labels, rather than a taxonomy of tree types.

\begin{figure*}[t]
\centering
\includegraphics[width=\textwidth]{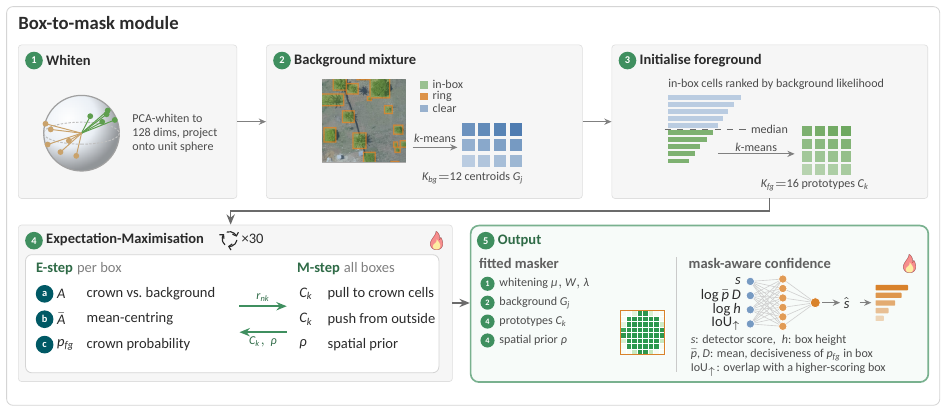}
\caption{Fitting the box-to-mask module once on training boxes: (1) whiten features, (2) fit a background mixture, (3) seed foreground prototypes, (4) learn prototypes and spatial prior by EM. At inference (Figure~\ref{fig:architecture}), the fitted masker is applied to each predicted box and (5) each prediction receives a mask-aware confidence.\label{fig:masker-fit}}
\end{figure*}

We observe that a crown typically occupies a roughly central, roughly circular region of a reasonably tight GT box. We encode this as a spatial prior $\rho$ over the crown's \emph{relative} position within the box, on an $N \times N$ grid of bins in normalised box coordinates $(u, v) \in [0,1]^2$, so $\rho$ is by construction invariant to crown size, tile size and ground sample distance. $\rho_k(b)$ is the prior that a cell in bin $b$ is crown and also explained by prototype $k$. $\hat\rho_k = \rho_k / \sum_{k'} \rho_{k'}$ normalises it over the $K_{fg}$ prototypes, while the unnormalised total $\sum_k \rho_k$ remains the probability that the bin is crown at all. $\rho$ is initialised flat, and at every M-step its mean per-bin foreground mass is rescaled to a nominal $\pi/4 \approx 0.785$, the area fraction of an ellipse inscribed in its bounding box for any aspect ratio (realised $0.690$ after the per-bin cap, Appendix~\ref{app:method}).

\medskip\noindent\emph{E-step.} For the cells of one box, with $\rho_k$ the spatial prior evaluated at each cell's bin for prototype $k$ and $\sum_k \rho_k$ the total foreground prior at that bin:

\begin{align}
A(z) &\;=\; \log \textstyle\sum_k \hat\rho_k \exp\!\left(\kappa\, z^{\top} C_k\right) \;-\; \ell_{bg}(z)
\label{eq:logratio}\\[2pt]
\bar A(z) &\;=\; A(z) - \operatorname{mean}_{z' \in \mathrm{box}} A(z')
\label{eq:meancentre}\\[2pt]
p_{fg}(z) &\;=\; \sigma\!\left(\bar A(z) \;+\; \alpha \, \mathrm{logit}\!\left(\textstyle\sum_k \rho_k\right)\right)
\label{eq:posterior}
\end{align}

Equation~\eqref{eq:logratio} is a log-likelihood ratio. Its first term is the foreground log-likelihood, built from the prototypes $C_k$ exactly as Equation~\eqref{eq:bgll} is built from the background centroids $G_j$; subtracting $\ell_{bg}$ gives $A(z) = \log\!\big(p(z \mid \text{crown}) / p(z \mid \text{background})\big)$. A cell must therefore look crown-like \emph{and} unlike background to score highly. This removes the soil/ground that resembles the background mixture outright.

Equation~\eqref{eq:meancentre} is a second filter. Attention over the surrounding tile inflates the absolute foreground score of \emph{every} cell inside a box, so $A$ is confounded by context and its absolute level is not sufficiently informative (Table~\ref{tab:components}): e.g.\ soil surrounded by canopy can still score positively, and therefore survive the first filter. However, mean-centring (subtracting the box mean of $A$ from each cell) isolates $A$ \emph{within} a box, asking not whether a cell is crown-like but whether it is \emph{more} crown-like than the rest of its own box. The behaviour that follows is as intended: a heterogeneous box containing crown and soil has a wide spread of $\bar A$ and is carved appropriately, whereas a box in uniformly closed canopy has a narrow spread and stays closer to filled, which is generally correct.

Alongside $p_{fg}$, the E-step returns a responsibility $r_{nk}$ for each cell $n$ of the box (with whitened embedding $z_n$) and each prototype $k$:

\begin{equation}
r_{nk} \;=\; p_{fg}(z_n)\;
\frac{\rho_k \exp\!\left(\kappa\, z_n^{\top} C_k\right)}
     {\sum_{k'} \rho_{k'} \exp\!\left(\kappa\, z_n^{\top} C_{k'}\right)}
\label{eq:resp}
\end{equation}

which together distribute each cell's foreground probability over the $K_{fg}$ prototypes in proportion to the weighted components of Equation~\eqref{eq:logratio}, so that $\sum_k r_{nk} = p_{fg}(z_n)$.

\medskip\noindent\emph{M-step.} The prototypes are updated in two steps. The first is the ordinary generative pull: prototype $k$ is drawn towards $P_k$, the unit-normalised responsibility-weighted mean $\sum_n r_{nk}\, z_n$ of all cells over all boxes, which expresses the idea ``recurs across boxes''. The second move repels the prototype from the outside-box signature it most resembles. The background cells are softmax-assigned to the current prototypes at the same concentration $\kappa$, with cells in a one-cell ring around each box weighted twice: they share illumination, phenology and local context (e.g.\ ground shadow) with the crown and so are the hardest negatives. The resulting per-prototype negative centroid $\bar N_k$ is the normalised weighted mean of the cells assigned to prototype $k$, and the prototype becomes

\begin{equation}
C_k \;=\; \frac{P_k - \beta\,\bar N_k}{\lVert P_k - \beta\,\bar N_k \rVert}
\label{eq:contrastive}
\end{equation}

A soil-like prototype's nearest negatives are themselves soil, so the subtraction repels it from soil: commonality is defined as agreement-across-boxes \emph{minus} outside-box signature, not merely whatever fills the box. The contrastive weight $\beta$ controls the strength of that repulsion; $\beta = 0$ recovers the purely generative update $C_k = P_k$ (compared in Table~\ref{tab:components}), and we use $\beta = 0.5$ throughout. The spatial prior is re-estimated from the same responsibilities: $\tilde\rho_k(b)$ is the mean of $r_{nk}$ over the cells $n$ falling in bin $b$, then constrained as described above to give $\rho$. Without this prior ($\alpha = 0$), test mask AP$_{50}$ falls by $0.022$ (Table~\ref{tab:masker}).

Fitting is deterministic given the seed. The module is fitted on the $8$~px lattice, a subset of the $4$~px lattice, and applied unchanged on the $4$~px lattice. We found that refitting on the $4$~px lattice, with $120$ or $250$ training tiles, only changes validation mask AP$_{50}$ by at most $0.001$, consistent with both fits sampling the same feature distribution.

\begin{algorithm}[t]
\caption{Box-to-mask fit (once, boxes only; $\kappa = 10$, $\beta = 0.5$, $T = 30$).\label{alg:fit}}
\begin{algorithmic}[1]
\State Whiten and normalise every cell's feature to a unit vector $z$ (Equation~\eqref{eq:whiten}).
\State Learn what background looks like: cluster outside-box cells into a background mixture and \textbf{freeze} it (Equation~\eqref{eq:bgll}).
\State Seed crown prototypes $C$ from in-box cells least like background; start the spatial prior $\rho$ flat.
\For{$t = 1$ \textbf{to} $T$}
  \State Score each in-box cell as crown or not ($p_{fg}$): more crown-like than background, relative to its own box, and in a likely position; share that score among the prototypes ($r_{nk}$) (Equations~\eqref{eq:logratio}--\eqref{eq:resp}). \Comment{E-step}
  \State Pull each prototype towards the in-box cells it explains, across all boxes, and push it away from the outside-box cells it most resembles (Equation~\eqref{eq:contrastive}). \Comment{M-step}
  \State Re-estimate where in a box crowns tend to sit ($\rho$), rescaling its total mass to $\pi/4$.
\EndFor
\State \Return whitening, frozen background, prototypes $C$, spatial prior $\rho$
\end{algorithmic}
\end{algorithm}

\subsection{Inference}
\label{sec:inference}

The posterior $p_{fg}$ of each predicted box lives on the $4$~px feature grid, so we upsample it to the native $2048$~px tile with a guided filter~\cite{guidedfilter} that uses the RGB tile as its guide, so that mask boundaries follow edges in the image rather than the feature grid. Pixels above a mask threshold $\tau$ form the predicted crown mask, which is intersected with the box, so a mask is contained in its box by construction rather than by penalty. $\tau$ is set per box to $0.33$ times the maximum of the filtered posterior inside the box, so that small crowns under-covered by imprecise predicted boxes are recovered. The prior weight $\alpha = 0.3$ scales the box-derived spatial term in Equation~\eqref{eq:posterior}: values $< 1$ relax reliance on a \emph{predicted} box, which is not always precise, so that the appearance term dominates. The concentration $\kappa$ is raised from $10$ at fitting to $16$ at inference in Equations~\eqref{eq:bgll} and~\eqref{eq:logratio}, sharpening the appearance term $\bar A$, which is zero-mean within each box.

\subsection{Confidence from the Mask Posterior}
\label{sec:scorer}

The detector score $s$ (Section~\ref{sec:detector}) rates whether a crown is present in the box, not how good the resulting mask (Section~\ref{sec:masker}) is. To generate a confidence score we can reduce the posterior over the $\lvert B \rvert$ cells of a box $B$ to two terms.

\begin{align}
\bar p(B) &\;=\; \frac{1}{\lvert B \rvert} \sum_{n \in B} p_{fg}(z_n)
\label{eq:meanpost}\\[2pt]
D(B) &\;=\; \frac{1}{\lvert B \rvert} \sum_{n \in B}
              \bigl\lvert\, 2\,p_{fg}(z_n) - 1 \,\bigr\rvert
\label{eq:decisive}
\end{align}

Because the mean-centring of Equation~\eqref{eq:meancentre} sets the mean of $\bar A$ to zero within every box, both statistics read the \emph{shape} of the within-box posterior rather than its level. Read this way $\bar p$ is a soft estimate of the fraction of its box that the crown occupies. The decisiveness $D$ folds each cell to its distance from $\tfrac12$ before averaging, and so records the \emph{spread} of the same distribution: $D = 1$ when every cell is confidently inside or outside, $D = 0$ when every cell is equally likely to be inside or outside.

A small multi-layer perceptron (MLP) gives the final mask-aware confidence $\hat s$. It takes four inputs per detection: $s$, $\log(\bar p\, D)$, the log box height $\log h$, and $\mathrm{IoU}_{\uparrow}$, the largest box IoU with any higher-scoring detection, which marks likely duplicates. It has one hidden layer of 64 units, trained on the validation detections to predict whether each matches a GT crown at box IoU $\geq 0.5$; detections falling in canopy are excluded, as they are in scoring. The deployed $\hat s$ is the mean of $25$ such MLPs ($5$ seeds $\times$ $5$ validation folds).

\subsection{Supervision and Model Selection}
\label{sec:labels}

No crown polygon is read at any stage of encoding, detection or mask fitting, and OAM-TCD's canopy class enters only as a region to ignore, as it does in scoring. Mask labels are used only on the $108$ validation images held out from the $900$ training images, to select the inference settings of Section~\ref{sec:inference} and the ranking of Section~\ref{sec:scorer} (full selection history in Appendix~\ref{app:selection}). This follows standard practice for box-supervised methods, whose hyperparameters are selected by mask AP on held-out masks (BoxInst on COCO val2017~\cite{boxinst}, Box2Mask on Pascal VOC val~\cite{box2mask}). The Box2Mask and BoxTeacher baselines used here inherit these settings from their released configurations. The LACE box-to-mask settings were evaluated once on the test set in a single pre-specified run.

\section{Results}
\label{sec:results}

\subsection{Datasets}
The OAM-TCD dataset~\cite{oamtcd}, introduced by Veitch-Michaelis et al., uses OpenAerialMap (OAM)~\cite{oam} VHR RGB imagery and provides $5{,}072$ $2048\times2048$~px images at $0.1$~m/px resolution with associated human-labelled instance masks for over $280$k individual and $56$k groups of trees (``canopy''). The OAM-TCD dataset provides polygon masks, as well as a broad range of ecological biomes, also including examples of the savanna \& dryland landscapes so scarce in existing annotated datasets. To our knowledge it is the only public dataset with individual crown masks spanning sparse-canopy biomes, and is therefore the only benchmark on which sparse-canopy TCIS can be measured.

Of the $5{,}072$ images in the OAM-TCD dataset, $439$ are reserved as a holdout benchmark test set, allowing for a global multi-biome comparison between models (Table~\ref{tab:test-biomes}, Appendix~\ref{app:datasets}). The $439$-image test set contains $25{,}705$ ITC annotations, and $366$ images within the set contain $\geq 1$ crown, i.e., $73$ of the images have $0$ ITCs. The same test set also contains $6{,}782$ canopy annotations, which are groups of trees in a wide range of real-world sizes. During creation of the dataset, human annotators were told to use the canopy class when it was not possible to reliably delineate individual trees, including when it was ``not obvious whether a tree was an individual or multiple''. The densest tile holds $422$ individual crowns, and the median tile holds $29$. Our $900$ training images are a random sample (seed $0$) of the OAM-TCD training tiles that contain at least one individual crown, split by tile into $792$ for training and $108$ for validation.

\subsection{Implementation Details}
\label{sec:implementation}

Table~\ref{tab:config} (Appendix~\ref{app:selection}) collects every key configuration used. The EM fit requires only a central processing unit (CPU) and completes in minutes. It is performed once, and the same fitted masker is used for every result reported.

\subsection{Tree Crown Instance Segmentation Results}
\label{sec:tcis}

For TCIS, we restrict our evaluation to OAM-TCD's Tree class (cat=2). One approach in the literature~\cite{selvabox}, which we adopt in this work, is to ignore all canopy annotations (cat=1) altogether. The model's predictions often fall inside areas marked as canopy, which contain entirely valid but merged tree crowns, but we exclude these from the evaluation metrics entirely (using COCO's iscrowd=1 to ignore). The rationale here is that the annotating of canopy into groups leads to subjective groupings. For example, should a clump of $10$ tree crowns be best delineated by $1$ canopy of $10$ crowns, or $2$ of $5$? We claim both are equally valid, and by using subjective groupings, metrics such as AP and F1 become unreliable since IoU thresholds are based on agreement of delineations. Every OAM-TCD figure we report is therefore ``canopy-neutral'', unless otherwise specified.

\begin{table*}[t]
\footnotesize\setlength{\tabcolsep}{3pt}
\caption{Instance segmentation on the OAM-TCD $439$-image holdout, all rows run and scored by us (COCOeval, native $2048^2$ tiles, single runs). Method names give the detection backbone; LACE's masker also reads frozen DINOv3 ViT-L/16 features. \emph{Imgs}: OAM-TCD training images (all models are pre-trained), rows with $900$ were trained by us. Configuration: Table~\ref{tab:geometry}. All LACE mask AP margins are significant (paired tile bootstrap, $2{,}000$ resamples, $95\%$ interval excludes zero). TP/FP/FN/F1: mask IoU $\geq 0.5$, each model at its own best-F1 score threshold, an oracle operating point applied equally to every row; the AP columns are threshold-free.}
\label{tab:oamtcd439}
\setlength{\tabcolsep}{4pt}
\begin{tabularx}{\linewidth}{@{}>{\raggedright\arraybackslash}X l r c c c r r r c@{}}
\toprule
 & \multicolumn{2}{c}{\textbf{Training}} & \multicolumn{2}{c}{\textbf{Mask}} & \textbf{Box} & \multicolumn{4}{c}{\textbf{Mask IoU $\mathbf{\geq 0.5}$}} \\
\cmidrule(lr){2-3} \cmidrule(lr){4-5} \cmidrule(lr){6-6} \cmidrule(l){7-10}
\textbf{Method} & \textbf{Superv.} & \textbf{Imgs} &
\textbf{AP\boldmath$_{50}$}\,$\uparrow$ &
\textbf{AP\boldmath$_{50:95}$}\,$\uparrow$ &
\textbf{AP\boldmath$_{50}$}\,$\uparrow$ &
\textbf{TP}\,$\uparrow$ & \textbf{FP}\,$\downarrow$ & \textbf{FN}\,$\downarrow$ & \textbf{F1}\,$\uparrow$ \\
\midrule
Mask R-CNN (OAM-TCD) & masks & $4{,}169$ & $0.648$ & $0.333$ & $0.665$ & $17{,}134$ & $6{,}983$ & $8{,}571$ & $0.688$ \\
Detectree2 R-101 & masks & $900$ & $0.630$ & $0.310$ & $0.651$ & $15{,}967$ & $8{,}766$ & $9{,}738$ & $0.633$ \\
\addlinespace[2pt]
Box2Mask Swin-L & box & $900$ & $0.589$ & $0.245$ & $0.620$ & $14{,}653$ & $\mathbf{5{,}460}$ & $11{,}052$ & $0.640$ \\
Box2Mask R-50 & box & $900$ & $0.451$ & $0.165$ & $0.512$ & $12{,}116$ & $6{,}930$ & $13{,}589$ & $0.542$ \\
BoxTeacher Swin-B & box & $900$ & $0.664$ & $0.305$ & $0.703$ & $16{,}218$ & $7{,}136$ & $9{,}487$ & $0.661$ \\
\addlinespace[2pt]
\textbf{LACE (ours)} Swin-L & box & $900$
  & $\mathbf{0.714}$ & $\mathbf{0.342}$ & $\mathbf{0.747}$ & $\mathbf{17{,}438}$ & $6{,}713$ & $\mathbf{8{,}267}$ & $\mathbf{0.700}$ \\
\bottomrule
\end{tabularx}
\end{table*}

Table~\ref{tab:oamtcd439} summarises our model's performance across COCO average precision (AP; with a single class, mAP and AP are the same) for OAM-TCD. Every baseline in Table~\ref{tab:oamtcd439} was re-run and re-scored by us under a single protocol (Appendix~\ref{app:protocol}), over all $439$ images including those with zero predictions. We use COCO defaults with three exceptions: maxDets $600$ in place of $100$, necessary since tiles in the test dataset contain up to 422 GT crowns; score floor $0.05$, inherited from Meta FAIR's Detectron2 test-time default setting~\cite{detectron2}; and canopy as \texttt{iscrowd} as discussed above.  Each baseline was run in its own native prediction geometry, so only the measurement is held fixed, and we comment in Table~\ref{tab:geometry} on any configuration choice that materially affects a row. Box2Mask and BoxTeacher were trained by us with identical supervision to LACE ($900$ images, boxes only), using their strongest released COCO box-supervised checkpoints, Swin-L and Swin-B backbones respectively (no Swin-L BoxTeacher has been released). Box2Mask reaches $0.589$ mask AP$_{50}$, and BoxTeacher reaches $0.664$ against LACE's $0.714$ under the same scorer. LACE is ahead of Mask R-CNN (OAM-TCD) at AP$_{50}$ ($0.714$ against $0.648$) and at box AP$_{50}$ ($0.747$ against $0.665$). The Mask R-CNN (OAM-TCD) score here is well above the published $0.432$ because our protocol is more favourable to it: restoring Detectron2's default proposal budget alone lifts it from $0.585$ to $0.648$ (Table~\ref{tab:rpn}). LACE's margin has two sources, both trained on boxes alone: a stronger detector (box AP$_{50}$ $0.747$ against $0.703$ for BoxTeacher) and the box-to-mask module, which we isolate in Section~\ref{sec:samebox}. Backbone size alone does not order the table: Box2Mask, with the same Swin-L backbone as LACE's detector, trails BoxTeacher's Swin-B by $0.075$. Because the module is decoupled from detection, it can mask any detector's boxes, and every gain in box detection carries over to masks with no further annotation (Section~\ref{sec:samebox}).

LACE also has the highest F1 ($0.700$), with the most true positives (TP) and fewest false negatives (FN; i.e. missed crowns), at the same IoU $\geq 0.5$ matching threshold~\cite{ball2023,weinstein2020,hao2021,plesoianu2020}. Box2Mask (Swin-L) produces the fewest false positives (FP), but is more cautious across the board and consequently produces significantly fewer true positives and more false negatives than other models, resulting in an F1 score of $0.640$.

\subsection{Annotation Efficiency Analysis}
\label{sec:cost}
\begin{figure*}[t]
\centering
\includegraphics[width=0.86\textwidth]{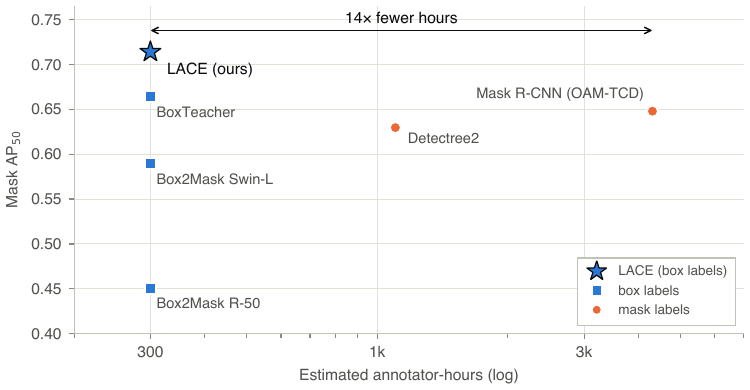}
\caption{Annotation cost against mask AP$_{50}$ on the OAM-TCD $439$-image holdout, one point per row of Table~\ref{tab:oamtcd439}. Hours are estimated from the OAM-TCD labels each row consumed in training or model selection ($54$~s per polygon, $7$~s per box~\cite{jain2013}, canopy groups charged as polygons throughout). Pretraining data are not counted, nor are the $108$ validation masks used once to fix LACE's inference settings.\label{fig:cost}}
\end{figure*}

Figure~\ref{fig:cost} sets each row of Table~\ref{tab:oamtcd439} against an estimate of the annotator time behind the OAM-TCD labels it consumed, whether in training or in model selection. We price a crown polygon at $54$~s and a box at $7$~s, the times measured on the same images and annotators in the user study of Jain and Grauman~\cite{jain2013}, and charge OAM-TCD's canopy groups as polygons for every row, since they are seen by every row. The $108$ validation masks that fixed LACE's inference settings and its ranking (Section~\ref{sec:labels}) are not charged here: like the baselines' own tuning, they are a one-off development cost, and we reuse the settings unchanged on all new data (the sparse-canopy, BoxTeacher detector, and PhenoBench tests below). On this estimate LACE consumes $300$ annotator-hours against $4{,}311$ for Mask R-CNN (OAM-TCD), $14$ times fewer, while scoring $0.066$ higher at AP$_{50}$. Other estimates put a polygon at up to $15$ times the annotation time cost of a box~\cite{khoreva2017}, and COCO's crowd workers took over $79$~s per polygon~\cite{coco}, which would widen the gap further.

\subsection{Sparse Canopy Segmentation Analysis}
The $439$-tile holdout holds $369$ forest tiles but only $16$ savanna and dryland tiles (Table~\ref{tab:test-biomes}), too few to measure on their own. To test model efficacy on sparse-canopy biomes such as savanna and dryland, we therefore take a `sparse-canopy' slice of the OAM-TCD data: $236$ tiles carrying $14{,}937$ crowns, none of which appears among the $900$ training images. We reuse models on this slice without any additional training or fine-tuning; Table~\ref{tab:sparse236} reports the result. To generate the slice, we categorise by the OAM-TCD dataset's own metadata which includes a \textit{biome} category for each image and select the most appropriate biomes by name, listed in Table~\ref{tab:sparse-biomes} (Appendix~\ref{app:datasets}). The slice is sparse by the yardstick of the Food and Agriculture Organization (FAO) as well as by name: labelled tree cover on its median tile is $7.5\%$, below the $10\%$ canopy cover at which FAO counts land as forest~\cite{fao2018terms}, and $59.3\%$ of its tiles fall under that line, compared with $31.0\%$ of the tiles drawn from the six forest biomes (labelled cover is a lower bound wherever annotators left gaps).

\begin{table*}[t]
\footnotesize\setlength{\tabcolsep}{3pt}
\caption{Instance segmentation on the sparse-canopy OAM-TCD slice ($236$ tiles, $14{,}937$ crowns, $220$ of the tiles from OAM-TCD's training split; no tile shared with our $900$ training images). N.B. Mask R-CNN (OAM-TCD) is omitted because it was trained on OAM-TCD's full training split, which contains most of these tiles. Canopy is scored as \texttt{iscrowd} throughout, as in Table~\ref{tab:oamtcd439}. All rows are single runs. All LACE mask AP margins are significant (paired tile bootstrap, as in Table~\ref{tab:oamtcd439}). TP/FP/FN/F1 as in Table~\ref{tab:oamtcd439}.\label{tab:sparse236}} \newcolumntype{C}{>{\centering\arraybackslash}X}
\setlength{\tabcolsep}{3pt}
\begin{tabularx}{\linewidth}{lCCCCCCC}
\toprule
\textbf{Method} & \textbf{AP\boldmath$_{50}$}\,$\uparrow$ &
\textbf{AP\boldmath$_{50:95}$}\,$\uparrow$ &
\textbf{Box AP\boldmath$_{50}$}\,$\uparrow$ &
\textbf{TP}\,$\uparrow$ & \textbf{FP}\,$\downarrow$ & \textbf{FN}\,$\downarrow$ & \textbf{F1}\,$\uparrow$ \\
\midrule
Detectree2 R-101 & $0.636$ & $0.353$ & $0.644$ & $9{,}172$ & $4{,}563$ & $5{,}765$ & $0.640$ \\
Box2Mask Swin-L & $0.604$ & $0.262$ & $0.627$ & $8{,}944$ & $3{,}380$ & $5{,}993$ & $0.656$ \\
BoxTeacher Swin-B & $0.721$ & $0.376$ & $0.739$ & $10{,}140$ & $3{,}013$ & $4{,}797$ & $0.722$ \\
\textbf{LACE (ours)} Swin-L & $\mathbf{0.743}$ & $\mathbf{0.403}$
  & $\mathbf{0.757}$ & $\mathbf{10{,}156}$ & $\mathbf{2{,}711}$ & $\mathbf{4{,}781}$ & $\mathbf{0.731}$ \\
\bottomrule
\end{tabularx}
\noindent{\footnotesize{Scored with the same frozen COCOeval protocol as Table~\ref{tab:oamtcd439} (pycocotools, masks at $2048^2$, maxDets $600$, canopy as \texttt{iscrowd}).}}
\end{table*}

\subsection{Visualisation Comparison}
\begin{figure*}[t]
\centering
\includegraphics[width=\textwidth]{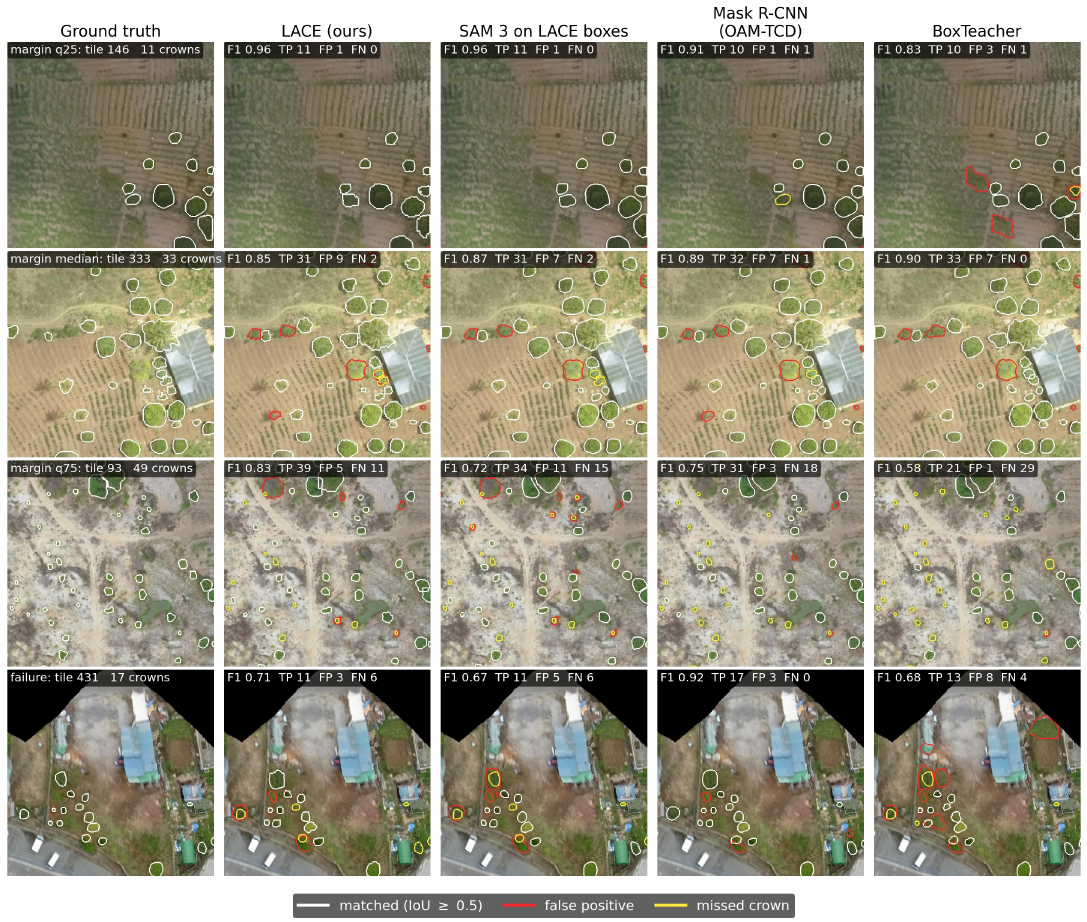}
\caption{Qualitative comparison on OAM-TCD holdout tiles ($512$~px crops, each model at its best-F1 threshold). Rows: open-ground tiles at the $25$th, $50$th and $75$th percentile of the LACE$-$Mask R-CNN (OAM-TCD) AP$_{50}$ margin, then a LACE failure case at the $5$th percentile.\label{fig:qualgrid}}
\end{figure*}

Figure~\ref{fig:qualgrid} compares the methods on three open-ground holdout tiles, together with a LACE failure case (SAM~3 on LACE's boxes is introduced in Section~\ref{sec:samebox}). Each paradigm is represented by its strongest Table~\ref{tab:oamtcd439} model, Mask R-CNN (OAM-TCD) for mask supervision and BoxTeacher for box supervision (Detectree2 and Box2Mask appear in Figure~\ref{fig:qual}). The three open-ground rows are the tiles closest to the $25$th, $50$th and $75$th percentiles of the LACE$-$Mask R-CNN (OAM-TCD) AP$_{50}$ margin. The failure row is the tile closest to the $5$th percentile; on it LACE leaves six crowns unmatched at IoU $0.5$, all of which Mask R-CNN (OAM-TCD) matches.

\subsection{Box-to-mask Analysis}
\label{sec:samebox}

To directly assess LACE's box-to-mask performance, we compare to controls that all share LACE's detector boxes (Table~\ref{tab:samebox}). The LACE masker beats ellipses by $+0.012$ AP$_{50}$ ($95\%$ paired tile-bootstrap interval $[0.010, 0.017]$) and $+0.024$ AP$_{50:95}$ ($[0.019, 0.029]$) on the test set, so its contribution is larger at stricter IoU thresholds. We also test SAM~3, used zero-shot as a drop-in masker, each ranked by its own score: LACE by $\hat s$, SAM~3 by $\sqrt{s\cdot s_{\mathrm{SAM}}}$, with $s_{\mathrm{SAM}}$ its predicted mask IoU, which outperformed using $s$ only (0.701 against 0.695 AP$_{50}$ on the test set). Each masker is thus ranked as its own pipeline would be deployed: $\hat s$ is part of LACE, and SAM~3 is given its strongest own score. On identical detector boxes, LACE is ahead at mask AP$_{50}$ on the test set ($0.714$ against $0.701$, a difference of $+0.013$ with $95\%$ paired tile-bootstrap interval $[0.007, 0.021]$). At a single operating point LACE also leads SAM~3 on F1 ($0.700$ against $0.684$), mainly through fewer false positives. SAM~3 leads at AP$_{50:95}$, so our box-fitted masker on frozen features matches a mask-pretrained foundation model at AP$_{50}$, but not at higher boundary precisions.

\begin{table*}[t]
\footnotesize\setlength{\tabcolsep}{3pt}
\caption{Same boxes, different maskers. Every row masks the identical set of LACE detections ($200{,}148$, OAM-TCD $439$). TP/FP/FN/F1 as in Table~\ref{tab:oamtcd439}.\label{tab:samebox}}
\newcolumntype{C}{>{\centering\arraybackslash}X}
\begin{tabularx}{\linewidth}{lCCCCCC}
\toprule
\textbf{Masker} & AP$_{50}$\,$\uparrow$ & AP$_{50:95}$\,$\uparrow$ & TP\,$\uparrow$ & FP\,$\downarrow$ & FN\,$\downarrow$ & F1\,$\uparrow$ \\
\midrule
\multicolumn{7}{l}{\textit{Shared ranking ($\hat s$ w/o mask input)}} \\
Filled box & $0.641$ & $0.217$ & $15{,}924$ & $8{,}161$ & $9{,}781$ & $0.640$ \\
Ellipse & $0.694$ & $0.316$ & $16{,}951$ & $7{,}076$ & $8{,}754$ & $0.682$ \\
LACE & $0.706$ & $0.340$ & $17{,}154$ & $6{,}852$ & $8{,}551$ & $0.690$ \\
\multicolumn{7}{l}{\textit{Own ranking}} \\
SAM~3 & $0.701$ & $\mathbf{0.355}$ & $17{,}190$ & $7{,}371$ & $8{,}515$ & $0.684$ \\
LACE & $\mathbf{0.714}$ & $0.342$ & $\mathbf{17{,}438}$ & $\mathbf{6{,}713}$ & $\mathbf{8{,}267}$ & $\mathbf{0.700}$ \\
\bottomrule
\end{tabularx}
\end{table*}

We also assess LACE's box-to-mask module unmodified on BoxTeacher's box detections. All configuration is frozen including $\hat s$ (the $s$ term used is BoxTeacher's detection box confidence score); no fine-tuning is applied anywhere. LACE's box-to-mask module beats BoxTeacher's mask head on the test set by $+0.013$ AP$_{50}$ ($[0.005, 0.021]$) and $+0.016$ AP$_{50:95}$ ($[0.011, 0.021]$) (Table~\ref{tab:btbox-test}). (Note that the reverse swap is not possible, as neither BoxTeacher's nor Box2Mask's mask head accepts an external box: each mask is tied to the method's own detection location or query~\cite{boxteacher,box2mask}).

\begin{table*}[t]
\footnotesize\setlength{\tabcolsep}{3pt}
\caption{Same boxes, different maskers. Every row masks the identical set of BoxTeacher detections ($263{,}132$, OAM-TCD $439$); LACE ranks them by $\hat s$ and BoxTeacher by its own score, so box AP$_{50}$ differs only through ranking. TP/FP/FN/F1 as in Table~\ref{tab:oamtcd439}. $\Delta$: LACE $-$ BoxTeacher, paired tile-bootstrap $95\%$ interval.\label{tab:btbox-test}}
\newcolumntype{C}{>{\centering\arraybackslash}X}
\setlength{\tabcolsep}{2pt}
\begin{tabularx}{\linewidth}{lCCCCCCCC}
\toprule
\textbf{Masker} & AP$_{50}$\,$\uparrow$ & AP$_{75}$\,$\uparrow$ & AP$_{50:95}$\,$\uparrow$ & Box AP$_{50}$\,$\uparrow$ & TP\,$\uparrow$ & FP\,$\downarrow$ & FN\,$\downarrow$ & F1\,$\uparrow$ \\
\midrule
BoxTeacher & $0.664$ & $0.223$ & $0.305$ & $0.703$ & $\mathbf{16{,}218}$ & $7{,}136$ & $\mathbf{9{,}487}$ & $0.661$ \\
LACE & $\mathbf{0.677}$ & $\mathbf{0.253}$ & $\mathbf{0.321}$ & $\mathbf{0.713}$ & $16{,}188$ & $\mathbf{6{,}651}$ & $9{,}517$ & $\mathbf{0.667}$ \\
\midrule
$\Delta$ [$95\%$ CI] & $+0.013$ & $+0.030$ & $+0.016$ & & & & & \\
 & \mbox{\scriptsize [$0.005$,\,$0.021$]} & \mbox{\scriptsize [$0.020$,\,$0.040$]} & \mbox{\scriptsize [$0.011$,\,$0.021$]} & & & & & \\
\bottomrule
\end{tabularx}
\end{table*}

\subsection{Zero-shot Transfer}
\label{sec:phenobench}

To assess transferability to out-of-domain datasets, we apply the LACE box-to-mask module completely unmodified to crop and weed plants in PhenoBench~\cite{phenobench}, with every masker prompted by the same GT boxes (Table~\ref{tab:phenobench}). It transfers, beating the inscribed ellipse over all $772$ images and the standard vegetation colour index. However, on this task SAM~3 zero-shot remains ahead at $0.719$ (Tables~\ref{tab:phenobench-protocol}--\ref{tab:phenobench-choice}).

\begin{table}[t]
\footnotesize\setlength{\tabcolsep}{3pt}
\caption{Zero-shot transfer of the LACE masker to PhenoBench crop and weed plants ($772$ validation images, $8{,}568$ plants), mean per-instance mask IoU from identical GT boxes.\label{tab:phenobench}}
\newcolumntype{C}{>{\centering\arraybackslash}X}
\begin{tabularx}{\linewidth}{Xcccc}
\toprule
\textbf{Masker} & \textbf{All} & \textbf{Crop} & \textbf{Weed} & \textbf{$\Delta$ LACE masker $-$ row} \\
\midrule
Filled box & $0.412$ & $0.410$ & $0.414$ & $+0.193$ {\scriptsize [$0.191$, $0.196$]} \\
Inscribed ellipse & $0.445$ & $0.440$ & $0.452$ & $+0.160$ {\scriptsize [$0.158$, $0.162$]} \\
Excess Green + Otsu~\cite{woebbecke1995} & $0.509$ & $0.624$ & $0.350$ & $+0.097$ {\scriptsize [$0.086$, $0.107$]} \\
LACE masker (tree-fitted) & $\mathbf{0.605}$ & $\mathbf{0.631}$ & $\mathbf{0.569}$ & \\
\bottomrule
\end{tabularx}
\end{table}

\subsection{Ablation Studies}
\label{sec:ablation}

We ablate LACE in pipeline order, from the interlaced feature grid to the box-to-mask module (E-step mean-centring, M-step repulsion and spatial prior) and its mask-aware confidence $\hat s$. Unless stated otherwise, ablations of the module hold the detector fixed, so every arm scores the identical set of predicted boxes and any difference is attributable to mask generation alone. All ablations use a single detector run. Test-split ablations report paired tile-bootstrap $95\%$ intervals, and the mean-centring and repulsion ablation is on the validation split. Further ablations of the module (dense canopy and foreground collapse) are given in Appendix~\ref{app:ablations}.

LACE interlaces offset passes of the frozen encoder into a real $4$~px grid (Section~\ref{sec:features}). With detections, module and scorer held fixed (Table~\ref{tab:grid}), interlacing beats interpolation on every metric, modestly at AP$_{50}$ ($+0.003$) and more at AP$_{75}$ ($+0.014$), where boundary placement matters most. Interpolation adds sampling resolution but no new evidence, whereas the offsets add evidence. Four offsets $\{0, p/2\}^2$ match sixteen at AP$_{50}$ ($0.000$, $[-0.001, 0.002]$) and trail by $0.001$ at AP$_{50:95}$. The gain from interlacing is therefore essentially entirely that of the first four offsets, which we propose as the default deployment configuration.

The box-to-mask module owes most of its accuracy to two components, E-step mean-centring and the spatial prior. Mean-centring (Equation~\eqref{eq:meancentre}) scores each cell against its own box, removing the background the box shares with its surroundings. It is the largest single effect, lifting mean crown IoU by $0.054$ on GT boxes and $0.059$ on LACE's predicted boxes (Table~\ref{tab:components}, validation split). M-step repulsion (Equation~\eqref{eq:contrastive}) removes the same background from the prototype side, a second route that lifts predicted-box crown IoU by $0.020$ on its own. Combined, the two give the best result on both GT and predicted boxes, although the repulsion adds only an additional $0.003$ and $0.0015$ respectively after mean-centring is applied.

\begin{table}[t]
\footnotesize\setlength{\tabcolsep}{3pt}
\caption{M-step repulsion against E-step mean-centring, $K_{fg}=16$, EM seed $0$, $108$ validation tiles, all on the deployed $4$~px path. ``Absolute'' disables mean-centring both when the masker is fitted and when it is applied. Mean per-crown mask IoU over $7{,}041$ crowns for the oracle-box column and $6{,}143$ box-matched true positives of the LACE detector for the predicted-box column.\label{tab:components}}
\newcolumntype{C}{>{\centering\arraybackslash}X}
\begin{tabularx}{\linewidth}{llCC}
\toprule
\textbf{M-step} & \textbf{E-step} & \textbf{GT Oracle boxes} & \textbf{Predicted boxes} \\
\midrule
generative ($\beta = 0$)     & absolute  & $0.7412$ & $0.6873$ \\
generative ($\beta = 0$)     & mean-centred & $0.7950$ & $0.7461$ \\
contrastive ($\beta = 0.5$)  & absolute  & $0.7585$ & $0.7075$ \\
contrastive ($\beta = 0.5$)  & mean-centred & $\mathbf{0.7980}$ & $\mathbf{0.7476}$ \\
\bottomrule
\end{tabularx}
\end{table}

The spatial prior is the second. Removing it costs $0.022$ AP$_{50}$ and $0.024$ AP$_{50:95}$ on the test set ($95\%$ intervals $[0.017, 0.028]$ and $[0.021, 0.027]$, Table~\ref{tab:masker}). Tuning $\alpha$ and $\kappa$ on validation adds a further $0.015$ AP$_{50}$ over the untuned setting ($0.699$ to $0.714$).

\begin{table}[t]
\footnotesize\setlength{\tabcolsep}{3pt}
\caption{Inference scalars on the OAM-TCD $439$-tile test split, scalars selected on $108$ held-out validation images. All rows use the same LACE detections ranked by the learned confidence $\hat s$ (Section~\ref{sec:scorer}), its frozen weights re-applied to each masker row. The comparison is paired. $\alpha$ and $\kappa$ were selected jointly; the middle row leaves both untuned (prior at full weight, $\kappa$ at its fitting value).\label{tab:masker}}
\newcolumntype{C}{>{\centering\arraybackslash}X}
\begin{adjustbox}{max width=\linewidth}\begin{tabular}{lcc}
\toprule
\textbf{Configuration} & \textbf{AP$_{50}$} & \textbf{AP$_{50:95}$} \\
\midrule
$\alpha = 0$ (no spatial prior), $\kappa = 16$ & $0.692$ & $0.318$ \\
$\alpha = 1.0$, $\kappa = 10$ & $0.699$ & $0.330$ \\
$\alpha = 0.3$, $\kappa = 16$ & $\mathbf{0.714}$ & $\mathbf{0.342}$ \\
\bottomrule
\end{tabular}\end{adjustbox}
\end{table}

The learned mask-aware confidence $\hat s$ adds $+0.024$ AP$_{50}$ over the raw detector confidence $s$ on the $439$ test tiles (Table~\ref{tab:factors}). It is fitted on validation detections only, was scored once on test, and transfers frozen to BoxTeacher's detections (Table~\ref{tab:btbox-test}).

\begin{table}[t]
\footnotesize\setlength{\tabcolsep}{3pt}
\caption{Mask-confidence scorer ablation for LACE. Every row ranks the same frozen set of $200{,}148$ test predictions with identical masks. Each ``$-$'' row refits it on the $108$ validation tiles without that input.}
\label{tab:factors}
\setlength{\tabcolsep}{3pt}
\begin{tabularx}{\linewidth}{@{}>{\raggedright\arraybackslash}X c c r r r c@{}}
\toprule
 & \multicolumn{6}{c}{\textbf{OAM-TCD $\mathbf{439}$}} \\
\cmidrule(l){2-7}
\textbf{Ranking} & \textbf{AP\boldmath$_{50}$}\,$\uparrow$ & \textbf{AP\boldmath$_{50:95}$}\,$\uparrow$ & \textbf{TP}\,$\uparrow$ & \textbf{FP}\,$\downarrow$ & \textbf{FN}\,$\downarrow$ & \textbf{F1}\,$\uparrow$ \\
\midrule
$s$ (detector score, mask-blind)            & $0.690$ & $0.329$ & $16{,}514$ & $6{,}781$ & $9{,}191$ & $0.674$ \\
\addlinespace[2pt]
$\hat s$ ($4$ inputs)           & $\mathbf{0.714}$ & $\mathbf{0.342}$ & $\mathbf{17{,}438}$ & $6{,}713$ & $\mathbf{8{,}267}$ & $\mathbf{0.700}$ \\
$\quad -\, s$                   & $0.641$ & $0.304$ & $15{,}531$ & $8{,}228$ & $10{,}174$ & $0.628$ \\
$\quad -\, \log(\bar p\,D)$     & $0.706$ & $0.340$ & $17{,}154$ & $6{,}852$ & $8{,}551$ & $0.690$ \\
$\quad -\, \log h$      & $0.707$ & $0.337$ & $16{,}878$ & $\mathbf{6{,}057}$ & $8{,}827$ & $0.694$ \\
$\quad -\, \mathrm{IoU}_{\uparrow}$       & $0.708$ & $0.341$ & $17{,}041$ & $6{,}502$ & $8{,}664$ & $0.692$ \\
\bottomrule
\end{tabularx}
\end{table}

\section{Discussion}
\label{sec:discussion}

\subsection{Box Supervision for Tree Crown Segmentation}
\label{sec:disc-box}

LACE's competitive results against other box-supervised models and fully supervised models promise a new viable option for TCIS tasks utilising time- and cost-efficient box annotations. It has demonstrated efficacy over a wide range of biomes, due to the flexibility of self-supervised embeddings to identify a diverse range of tree crowns, and it leads every model tested on the sparse-canopy slice (Table~\ref{tab:sparse236}), where the focus is on measuring tree crown count, size, and allometries over large, noisy landscapes. The foreground collapse (Table~\ref{tab:collapse}) is consistent with this. Fitted on boxes alone, the module captures crowns across OAM-TCD's biomes with a single direction in frozen DINOv3 features. 

One important implication of our results is that TCIS on OAM-TCD is a detection-led task, and we posit that it would be reasonable to assume this to be true of sparse-canopy environments generally. LACE's lead over Mask R-CNN (OAM-TCD) is larger at box than at mask AP$_{50}$ ($+0.082$ against $+0.066$), and we find that even an inscribed ellipse on LACE's boxes ($0.694$, Table~\ref{tab:samebox}) is above every baseline in Table~\ref{tab:oamtcd439}. What remains post-detection is to recover the crown reliably over boxes that can have varying degrees of imprecision, and LACE's box-to-mask module does so: it beats the inscribed ellipse on LACE's boxes (AP$_{50}$ $+0.012$, $[0.010, 0.017]$; F1 $0.690$ against $0.682$), the mask-pretrained foundation model SAM~3 on the same boxes (AP$_{50}$ $+0.013$, $[0.007, 0.021]$; F1 $0.700$ against $0.684$), and BoxTeacher's own mask head on BoxTeacher's boxes (AP$_{50}$ $+0.013$, $[0.005, 0.021]$; F1 $0.667$ against $0.661$).

\subsection{Limitations}
\label{sec:limitations}

Box-supervised models have an additional subtle disadvantage to their mask-supervised counterparts. Human annotations function as the ground truth, though strictly they are reference data and subject to error~\cite{foody2010,olofsson2014}, and therefore any manually annotated dataset naturally contains idiosyncrasies and a degree of interpretation in its annotations. This is particularly the case with respect to edge cases and ambiguities. Mask-supervised models are able to learn these annotation conventions to some degree, whereas box-supervised models are measured by an unseen yardstick. They are fully reliant on agreement between the actual real-world ground truth and the dataset's version of it, having not had any opportunity to adjust to the reference data's biases. This reliance is most limiting in dense canopy, where reference data often attempt definitive delineations that the image cannot support~\cite{ball2023,steier2024}.

By taking the $4$~px interlaced grid as the unit of crown delineation, LACE has a disadvantage relative to pixel-based maskers: its posterior is defined on a grid four times coarser than the image. The guided upsampling of Section~\ref{sec:inference} compensates for this at the native raster by letting the image decide where a boundary falls between cells. On the test set it adds $+0.026$ AP$_{75}$ and $+0.012$ AP$_{50:95}$ over bilinear upsampling (AP$_{50}$ $+0.004$), while a blur with no image guide and the RGB of a different tile both fall below bilinear on every metric, so the gain comes from the image's own edges rather than from smoothing. LACE leads the Mask R-CNN on AP$_{50:95}$ ($0.342$ against $0.333$) but still trails it at certain strict IoUs (e.g.\ AP$_{75}$ $0.271$ against $0.297$). This is a key trade-off of using a masker that has not been trained on polygon masks. Further exploration of new boundary-aware decoder modules and approaches~\cite{zhang2024boundary, bare} would be a natural next step to improve performance at strict IoU configurations.

Although the spatial prior is maintained separately for three crown-size bands (Appendix~\ref{app:method}), in practice we observe very little difference between the three fitted priors, contrary to our expectations. We leave the stratification in as a defensive component that does not add any significant complexity, but acknowledge that a single pooled prior would likely perform comparably.

\subsection{Ecological Implications}
\label{sec:disc-broader}

In a warming world, it is tempting to reduce the objective to something along the lines of `tree maximalism', i.e.\ the position that more trees is always better. Open African savanna in particular, as argued by Bond et al.~\cite{bond2019}, has often been considered as a kind of deforested or degraded landscape~\cite{laestadius2012,bastin2019} when in fact the biodiversity~\cite{veldman2015,jenkins2013}, surface temperature through biophysical effects~\cite{bright2015}, streamflow~\cite{jackson2005}, and vulnerability to fire~\cite{veldman2015} are all negatively impacted by aggressive tree \emph{planting} in these environments, and moreover Dass et al.~\cite{dass2018} have argued that grasslands (in California) may actually be more reliable and robust carbon sinks than dense forests. Further analysis of this debate is beyond the scope of this work, but we call attention to the nuance required in practical implementation of tree crown delineation and carbon accounting using techniques such as LACE.

\section{Conclusions}
\label{sec:conclusions}

In this study we present LACE, an instance segmentation model in box-supervised tree crown tasks where crown polygon annotations are scarce. Our box-to-mask inference method provides a SOTA-competitive solution to true TCIS without the need for time-consuming and/or expensive polygon masks by leveraging the statistical signature of the frozen DINOv3 features. With this contribution, we hope to support the Remote Sensing community in the previously underexplored area of TCIS in diverse global biomes, particularly sparse-canopy landscapes. 

Our end-to-end evaluation uses OAM-TCD, currently the only multi-biome tree crown mask benchmark, at $0.1$~m/px RGB, and extending LACE to coarser GSD and to multispectral, height (DSM, CHM) or other multimodal inputs such as LiDAR is left to future work. Future encoders with native per-pixel features would remove the lattice and guided-filter steps entirely, and the further exploration of the balance between embedding dimension, feature encoding, and spatial resolution in decoder-free maskers such as LACE's also warrants investigation.

\section*{Author Contributions}
Conceptualisation, methodology, software, validation, formal analysis, investigation, visualisation and writing---original draft preparation, T.P.; supervision, B.L.; writing---review and editing, K.L. and B.L. All authors have read and agreed to the published version of the manuscript.

\section*{Funding}
This research received no external funding.

\section*{Data Availability}
The dataset used in this study is publicly available: the OAM-TCD dataset~\cite{oamtcd} (\url{https://huggingface.co/datasets/restor/tcd}). All aerial imagery shown in the figures is from OAM-TCD, contributed by OpenAerialMap users to the Open Imagery Network under the CC BY 4.0 licence (\url{https://creativecommons.org/licenses/by/4.0/}), and has been cropped and overlaid with annotations and model predictions. All code for LACE, together with the training configurations, model checkpoints and evaluation scripts needed to reproduce every table in this paper, will be released upon publication at \url{https://github.com/tomdpitts/LACE}.

\section*{Conflicts of Interest}
The authors declare no conflicts of interest.

\section*{Use of AI Assistance}
During the preparation of this study, the authors used a large language model coding assistant, Anthropic's Claude Code (Claude Opus 5.4 and 5.5), for the purposes of writing and debugging the LACE implementation and evaluation code, launching and logging experiments, drafting and revising short passages of the manuscript, and rendering graphics. The research questions, method concept and design, choice of datasets and baselines, experiment design, graphic design \& layout, the manuscript text, all commentary, and interpretation of results are the authors' own; the assistant was used by the first author alone, who directed and checked every experiment from which the reported numbers come. The authors have reviewed and edited the output and take full responsibility for the content of this publication.

{
\small
\bibliographystyle{ieeenat_fullname}
\bibliography{lace}

\begin{thebibliography}{99}
\providecommand{\natexlab}[1]{#1}
\providecommand{\url}[1]{\texttt{#1}}
\expandafter\ifx\csname urlstyle\endcsname\relax
  \providecommand{\doi}[1]{doi: #1}\else
  \providecommand{\doi}{doi: \begingroup \urlstyle{rm}\Url}\fi

\bibitem[Ahlstr\"{o}m et~al.(2015)Ahlstr\"{o}m, Raupach, Schurgers, Smith,
  Arneth, Jung, Reichstein, Canadell, Friedlingstein, Jain, Kato, Poulter,
  Sitch, Stocker, Viovy, Wang, Wiltshire, Zaehle, and Zeng]{ahlstrom2015}
Anders Ahlstr\"{o}m, Michael~R. Raupach, Guy Schurgers, Benjamin Smith, Almut
  Arneth, Martin Jung, Markus Reichstein, Josep~G. Canadell, Pierre
  Friedlingstein, Atul~K. Jain, Etsushi Kato, Benjamin Poulter, Stephen Sitch,
  Benjamin~D. Stocker, Nicolas Viovy, Ying~Ping Wang, Andy Wiltshire, S\"{o}nke
  Zaehle, and Ning Zeng.
\newblock The dominant role of semi-arid ecosystems in the trend and
  variability of the land {CO$_2$} sink.
\newblock \emph{Science}, 348\penalty0 (6237):\penalty0 895--899, 2015.

\bibitem[Avitabile et~al.(2016)Avitabile, Herold, Heuvelink, Lewis, Phillips,
  Asner, Armston, Ashton, Banin, Bayol, et~al.]{avitabile2016}
Valerio Avitabile, Martin Herold, Gerard B.~M. Heuvelink, Simon~L. Lewis,
  Oliver~L. Phillips, Gregory~P. Asner, John Armston, Peter~S. Ashton, Lindsay
  Banin, Nicolas Bayol, et~al.
\newblock An integrated pan-tropical biomass map using multiple reference
  datasets.
\newblock \emph{Glob. Change Biol.}, 22\penalty0 (4):\penalty0 1406--1420,
  2016.

\bibitem[Baccini et~al.(2012)Baccini, Goetz, Walker, Laporte, Sun,
  Sulla-Menashe, Hackler, Beck, Dubayah, Friedl, Samanta, and
  Houghton]{baccini2012}
A. Baccini, S.~J. Goetz, W.~S. Walker, N.~T. Laporte, M. Sun, D. Sulla-Menashe,
  J. Hackler, P.~S.~A. Beck, R. Dubayah, M.~A. Friedl, S. Samanta, and R.~A.
  Houghton.
\newblock Estimated carbon dioxide emissions from tropical deforestation
  improved by carbon-density maps.
\newblock \emph{Nat. Clim. Change}, 2\penalty0 (3):\penalty0 182--185, 2012.

\bibitem[Ball et~al.(2023)Ball, Hickman, Jackson, Koay, Hirst, Jay, Archer,
  Aubry-Kientz, Vincent, and Coomes]{ball2023}
James G.~C. Ball, Sebastian H.~M. Hickman, Tobias~D. Jackson, Xian~Jing Koay,
  James Hirst, William Jay, Matthew Archer, M\'{e}laine Aubry-Kientz,
  Gr\'{e}goire Vincent, and David~A. Coomes.
\newblock Accurate delineation of individual tree crowns in tropical forests
  from aerial {RGB} imagery using {Mask R-CNN}.
\newblock \emph{Remote Sens. Ecol. Conserv.}, 9\penalty0 (5):\penalty0
  641--655, 2023.

\bibitem[Banerjee et~al.(2005)Banerjee, Dhillon, Ghosh, and
  Sra]{banerjee2005vmf}
Arindam Banerjee, Inderjit~S. Dhillon, Joydeep Ghosh, and Suvrit Sra.
\newblock Clustering on the unit hypersphere using von {M}ises--{F}isher
  distributions.
\newblock \emph{Journal of Machine Learning Research}, 6:\penalty0 1345--1382,
  2005.

\bibitem[Bastin et~al.(2017)Bastin, Berrahmouni, Grainger, Maniatis, Mollicone,
  Moore, Patriarca, Picard, Sparrow, Abraham, et~al.]{bastin2017}
Jean-Fran\c{c}ois Bastin, Nora Berrahmouni, Alan Grainger, Danae Maniatis,
  Danilo Mollicone, Rebecca Moore, Chiara Patriarca, Nicolas Picard, Ben
  Sparrow, Elena~Maria Abraham, et~al.
\newblock The extent of forest in dryland biomes.
\newblock \emph{Science}, 356\penalty0 (6338):\penalty0 635--638, 2017.

\bibitem[Bastin et~al.(2019)Bastin, Finegold, Garcia, Mollicone, Rezende,
  Routh, Zohner, and Crowther]{bastin2019}
Jean-Fran{\c{c}}ois Bastin, Yelena Finegold, Claude Garcia, Danilo Mollicone,
  Marcelo Rezende, Devin Routh, Constantin~M. Zohner, and Thomas~W. Crowther.
\newblock The global tree restoration potential.
\newblock \emph{Science}, 365\penalty0 (6448):\penalty0 76--79, 2019.

\bibitem[Baudchon et~al.(2026)Baudchon, Ouaknine, Weiss, Teng, Walla,
  Caron-Guay, Pal, and Lalibert\'{e}]{selvabox}
Hugo Baudchon, Arthur Ouaknine, Martin Weiss, M\'{e}lisande Teng, Thomas~R.
  Walla, Antoine Caron-Guay, Christopher Pal, and Etienne Lalibert\'{e}.
\newblock {SelvaBox}: A high-resolution dataset for tropical tree crown
  detection.
\newblock In \emph{14th International Conference on Learning Representations
  (ICLR 2026)}, 2026.
\newblock arXiv:2507.00170.

\bibitem[Bond et~al.(2019)Bond, Stevens, Midgley, and Lehmann]{bond2019}
William~J. Bond, Nicola Stevens, Guy~F. Midgley, and Caroline E.~R. Lehmann.
\newblock The trouble with trees: {Afforestation} plans for {Africa}.
\newblock \emph{Trends Ecol. Evol.}, 34\penalty0 (11):\penalty0 963--965, 2019.

\bibitem[Bouvet et~al.(2018)Bouvet, Mermoz, Le~Toan, Villard, Mathieu, Naidoo,
  and Asner]{bouvet2018}
Alexandre Bouvet, St\'{e}phane Mermoz, Thuy Le~Toan, Ludovic Villard, Renaud
  Mathieu, Laven Naidoo, and Gregory~P. Asner.
\newblock An above-ground biomass map of {African} savannahs and woodlands at
  25~m resolution derived from {ALOS PALSAR}.
\newblock \emph{Remote Sens. Environ.}, 206:\penalty0 156--173, 2018.

\bibitem[Braga et~al.(2020)Braga, Peripato, Dalagnol, Ferreira, Tarabalka,
  Arag\~{a}o, Campos~Velho, Shiguemori, and Wagner]{braga2020}
Jos\'{e} R.~G. Braga, Vin\'{i}cius Peripato, Ricardo Dalagnol, Matheus~P.
  Ferreira, Yuliya Tarabalka, Luiz E. O.~C. Arag\~{a}o, Haroldo F.~de
  Campos~Velho, Elcio~H. Shiguemori, and Fabien~H. Wagner.
\newblock Tree crown delineation algorithm based on a convolutional neural
  network.
\newblock \emph{Remote Sens.}, 12\penalty0 (8):\penalty0 1288, 2020.

\bibitem[Brandt and Stolle(2021)]{brandtstolle2021}
John Brandt and Fred Stolle.
\newblock A global method to identify trees outside of closed-canopy forests
  with medium-resolution satellite imagery.
\newblock \emph{Int. J. Remote Sens.}, 42\penalty0 (5):\penalty0 1713--1737,
  2021.

\bibitem[Brandt et~al.(2020)Brandt, Tucker, Kariryaa, Rasmussen, Abel, Small,
  Chave, Rasmussen, Hiernaux, Diouf, Kergoat, Mertz, Igel, Gieseke,
  Sch\"{o}ning, Li, Melocik, Meyer, Sinno, Romero, Glennie, Montagu,
  Dendoncker, and Fensholt]{brandt2020}
Martin Brandt, Compton~J. Tucker, Ankit Kariryaa, Kjeld Rasmussen, Christin
  Abel, Jennifer Small, Jerome Chave, Laura~Vang Rasmussen, Pierre Hiernaux,
  Abdoul~Aziz Diouf, Laurent Kergoat, Ole Mertz, Christian Igel, Fabian
  Gieseke, Johannes Sch\"{o}ning, Sizhuo Li, Katherine Melocik, Jesse Meyer,
  Scott Sinno, Eric Romero, Erin Glennie, Amandine Montagu, Morgane Dendoncker,
  and Rasmus Fensholt.
\newblock An unexpectedly large count of trees in the {West African Sahara} and
  {Sahel}.
\newblock \emph{Nature}, 587\penalty0 (7832):\penalty0 78--82, 2020.

\bibitem[Bright et~al.(2015)Bright, Zhao, Jackson, and Cherubini]{bright2015}
Ryan~M. Bright, Kaiguang Zhao, Robert~B. Jackson, and Francesco Cherubini.
\newblock Quantifying surface albedo and other direct biogeophysical climate
  forcings of forestry activities.
\newblock \emph{Glob. Change Biol.}, 21\penalty0 (9):\penalty0 3246--3266,
  2015.

\bibitem[Carion et~al.(2025)Carion, Gustafson, Hu, Debnath, Hu, Suris, Ryali,
  Alwala, Khedr, Huang, et~al.]{sam3}
Nicolas Carion, Laura Gustafson, Yuan-Ting Hu, Shoubhik Debnath, Ronghang Hu,
  Didac Suris, Chaitanya Ryali, Kalyan~Vasudev Alwala, Haitham Khedr, Andrew
  Huang, et~al.
\newblock {SAM 3}: Segment anything with concepts.
\newblock \emph{arXiv}, 2025.
\newblock arXiv:2511.16719.

\bibitem[Caron et~al.(2021)Caron, Touvron, Misra, J\'{e}gou, Mairal,
  Bojanowski, and Joulin]{dino}
Mathilde Caron, Hugo Touvron, Ishan Misra, Herv\'{e} J\'{e}gou, Julien Mairal,
  Piotr Bojanowski, and Armand Joulin.
\newblock Emerging properties in self-supervised vision transformers.
\newblock In \emph{IEEE/CVF International Conference on Computer Vision
  (ICCV)}, pages 9630--9640, 2021.

\bibitem[Chadwick et~al.(2020)Chadwick, Goodbody, Coops, Hervieux, Bater,
  Martens, White, and R\"{o}eser]{chadwick2020}
Andrew~J. Chadwick, Tristan R.~H. Goodbody, Nicholas~C. Coops, Anne Hervieux,
  Christopher~W. Bater, Lee~A. Martens, Barry White, and Dominik R\"{o}eser.
\newblock Automatic delineation and height measurement of regenerating conifer
  crowns under leaf-off conditions using {UAV} imagery.
\newblock \emph{Remote Sens.}, 12\penalty0 (24):\penalty0 4104, 2020.

\bibitem[Chen et~al.(2024)Chen, Liu, Chen, Zhang, Li, Zou, and Shi]{rsprompter}
Keyan Chen, Chenyang Liu, Hao Chen, Haotian Zhang, Wenyuan Li, Zhengxia Zou,
  and Zhenwei Shi.
\newblock {RSPrompter}: Learning to prompt for remote sensing instance
  segmentation based on visual foundation model.
\newblock \emph{IEEE Trans. Geosci. Remote Sens.}, 62:\penalty0 1--17, 2024.

\bibitem[Chen et~al.(2025{\natexlab{a}})Chen, Kong, Sun, Jiao, Zhao, and
  Liu]{ginkgo}
Mengyuan Chen, Wenwen Kong, Yongqi Sun, Jie Jiao, Yunpeng Zhao, and Fei Liu.
\newblock A two-stage segment-then-classify strategy for accurate ginkgo tree
  identification from {UAV} imagery.
\newblock \emph{Drones}, 9\penalty0 (11):\penalty0 773, 2025{\natexlab{a}}.

\bibitem[Chen et~al.(2025{\natexlab{b}})Chen, Russell, Pallavoor, Young, and
  Wu]{chen2025zeroshot}
Michelle Chen, David Russell, Amritha Pallavoor, Derek Young, and Jane Wu.
\newblock Zero-shot tree detection and segmentation from aerial forest imagery.
\newblock \emph{arXiv}, 2025{\natexlab{b}}.
\newblock arXiv:2506.03114.

\bibitem[Chen et~al.(2026)Chen, Lyu, Wang, and Wang]{fgtreeseg}
Pengyu Chen, Fangzheng Lyu, Sicheng Wang, and Cuizhen Wang.
\newblock {FG-TreeSeg}: Flow-guided tree crown segmentation without instance
  annotations.
\newblock \emph{IEEE Geosci. Remote Sens. Lett.}, 23:\penalty0 2503705, 2026.

\bibitem[Cheng et~al.(2023)Cheng, Wang, Chen, Zhang, and Liu]{boxteacher}
Tianheng Cheng, Xinggang Wang, Shaoyu Chen, Qian Zhang, and Wenyu Liu.
\newblock {BoxTeacher}: Exploring high-quality pseudo labels for weakly
  supervised instance segmentation.
\newblock In \emph{IEEE/CVF Conference on Computer Vision and Pattern
  Recognition (CVPR)}, pages 3145--3154, 2023.

\bibitem[Crowther et~al.(2015)Crowther, Glick, Covey, Bettigole, Maynard,
  Thomas, Smith, Hintler, Duguid, Amatulli, Tuanmu, Jetz, Salas, Stam, Piotto,
  Tavani, Green, Bruce, Williams, Wiser, Huber, Hengeveld, Nabuurs, Tikhonova,
  Borchardt, Li, Powrie, Fischer, Hemp, Homeier, Cho, Vibrans, Umunay, Piao,
  Rowe, Ashton, Crane, and Bradford]{crowther2015}
T.~W. Crowther, H.~B. Glick, K.~R. Covey, C. Bettigole, D.~S. Maynard, S.~M.
  Thomas, J.~R. Smith, G. Hintler, M.~C. Duguid, G. Amatulli, M.-N. Tuanmu, W.
  Jetz, C. Salas, C. Stam, D. Piotto, R. Tavani, S. Green, G. Bruce, S.~J.
  Williams, S.~K. Wiser, M.~O. Huber, G.~M. Hengeveld, G.-J. Nabuurs, E.
  Tikhonova, P. Borchardt, C.-F. Li, L.~W. Powrie, M. Fischer, A. Hemp, J.
  Homeier, P. Cho, A.~C. Vibrans, P.~M. Umunay, S.~L. Piao, C.~W. Rowe, M.~S.
  Ashton, P.~R. Crane, and M.~A. Bradford.
\newblock Mapping tree density at a global scale.
\newblock \emph{Nature}, 525\penalty0 (7568):\penalty0 201--205, 2015.

\bibitem[Dass et~al.(2018)Dass, Houlton, Wang, and Warlind]{dass2018}
Pawlok Dass, Benjamin~Z. Houlton, Yingping Wang, and David Warlind.
\newblock Grasslands may be more reliable carbon sinks than forests in
  {California}.
\newblock \emph{Environ. Res. Lett.}, 13\penalty0 (7):\penalty0 074027, 2018.

\bibitem[Deng et~al.(2009)Deng, Dong, Socher, Li, Li, and Fei-Fei]{imagenet}
Jia Deng, Wei Dong, Richard Socher, Li-Jia Li, Kai Li, and Li Fei-Fei.
\newblock {ImageNet}: A large-scale hierarchical image database.
\newblock In \emph{IEEE Conference on Computer Vision and Pattern Recognition
  (CVPR)}, pages 248--255, 2009.

\bibitem[Do et~al.(2026)Do, Phung, Pham, Pham, Hoang, Hoang, Vlaminck, Luong,
  Tran, Vu, and Le]{treecog}
Mai Viet~Hoang Do, Duc-Thang Phung, Hoang Duy~Linh Pham, Quang-Duy Pham,
  Van-Nam Hoang, Van-Sam Hoang, Michiel Vlaminck, Hiep Luong, Thanh-Hai Tran,
  Hai Vu, and Thi-Lan Le.
\newblock A {UAV} {RGB} dataset and method for instance tree crown segmentation
  for biodiversity monitoring.
\newblock \emph{Sci. Rep.}, 16:\penalty0 5788, 2026.

\bibitem[Dosovitskiy et~al.(2021)Dosovitskiy, Beyer, Kolesnikov, Weissenborn,
  Zhai, Unterthiner, Dehghani, Minderer, Heigold, Gelly, Uszkoreit, and
  Houlsby]{vit}
Alexey Dosovitskiy, Lucas Beyer, Alexander Kolesnikov, Dirk Weissenborn,
  Xiaohua Zhai, Thomas Unterthiner, Mostafa Dehghani, Matthias Minderer, Georg
  Heigold, Sylvain Gelly, Jakob Uszkoreit, and Neil Houlsby.
\newblock An image is worth 16x16 words: Transformers for image recognition at
  scale.
\newblock In \emph{9th International Conference on Learning Representations
  (ICLR)}, 2021.
\newblock arXiv:2010.11929.

\bibitem[Drusch et~al.(2012)Drusch, Del~Bello, Carlier, Colin, Fernandez,
  Gascon, Hoersch, Isola, Laberinti, Martimort, Meygret, Spoto, Sy, Marchese,
  and Bargellini]{drusch2012}
M. Drusch, U. Del~Bello, S. Carlier, O. Colin, V. Fernandez, F. Gascon, B.
  Hoersch, C. Isola, P. Laberinti, P. Martimort, A. Meygret, F. Spoto, O. Sy,
  F. Marchese, and P. Bargellini.
\newblock {Sentinel-2}: {ESA}'s optical high-resolution mission for {GMES}
  operational services.
\newblock \emph{Remote Sensing of Environment}, 120:\penalty0 25--36, 2012.

\bibitem[Duguay et~al.(2026)Duguay, Baudchon, Lalibert\'{e}, Muller-Landau,
  Rivas-Torres, and Ouaknine]{selvamask}
Simon-Olivier Duguay, Hugo Baudchon, Etienne Lalibert\'{e}, Helene
  Muller-Landau, Gonzalo Rivas-Torres, and Arthur Ouaknine.
\newblock {SelvaMask}: Segmenting trees in tropical forests and beyond.
\newblock \emph{arXiv}, 2026.
\newblock arXiv:2602.02426.

\bibitem[Fagan(2020)]{fagan2020}
Matthew~E. Fagan.
\newblock A lesson unlearned? {Underestimating} tree cover in drylands biases
  global restoration maps.
\newblock \emph{Glob. Change Biol.}, 26\penalty0 (9):\penalty0 4679--4690,
  2020.

\bibitem[{FAO}(2018)]{fao2018terms}
{FAO}.
\newblock Global forest resources assessment 2020: Terms and definitions.
\newblock Technical Report Forest Resources Assessment Working Paper 188, Food
  and Agriculture Organization of the United Nations, Rome, Italy, 2018.
\newblock \url{https://www.fao.org/3/I8661EN/i8661en.pdf} (accessed on 17
  September 2026).

\bibitem[Foody(2010)]{foody2010}
Giles~M. Foody.
\newblock Assessing the accuracy of land cover change with imperfect ground
  reference data.
\newblock \emph{Remote Sensing of Environment}, 114\penalty0 (10):\penalty0
  2271--2285, 2010.

\bibitem[Ghasemi et~al.(2022)Ghasemi, Latifi, and Pourhashemi]{ghasemi2022}
Marziye Ghasemi, Hooman Latifi, and Mehdi Pourhashemi.
\newblock A novel method for detecting and delineating coppice trees in {UAV}
  images to monitor tree decline.
\newblock \emph{Remote Sensing}, 14\penalty0 (23):\penalty0 5910, 2022.

\bibitem[Girshick et~al.(2014)Girshick, Donahue, Darrell, and Malik]{rcnn}
Ross Girshick, Jeff Donahue, Trevor Darrell, and Jitendra Malik.
\newblock Rich feature hierarchies for accurate object detection and semantic
  segmentation.
\newblock In \emph{IEEE Conference on Computer Vision and Pattern Recognition
  (CVPR)}, pages 580--587, 2014.

\bibitem[Hamilton et~al.(2022)Hamilton, Zhang, Hariharan, Snavely, and
  Freeman]{stego}
Mark Hamilton, Zhoutong Zhang, Bharath Hariharan, Noah Snavely, and William~T.
  Freeman.
\newblock Unsupervised semantic segmentation by distilling feature
  correspondences.
\newblock In \emph{10th International Conference on Learning Representations
  (ICLR)}, 2022.
\newblock arXiv:2203.08414.

\bibitem[Hansen et~al.(2013)Hansen, Potapov, Moore, Hancher, Turubanova,
  Tyukavina, Thau, Stehman, Goetz, Loveland, Kommareddy, Egorov, Chini,
  Justice, and Townshend]{hansen2013}
M.~C. Hansen, P.~V. Potapov, R. Moore, M. Hancher, S.~A. Turubanova, A.
  Tyukavina, D. Thau, S.~V. Stehman, S.~J. Goetz, T.~R. Loveland, A.
  Kommareddy, A. Egorov, L. Chini, C.~O. Justice, and J.~R.~G. Townshend.
\newblock High-resolution global maps of 21st-century forest cover change.
\newblock \emph{Science}, 342\penalty0 (6160):\penalty0 850--853, 2013.

\bibitem[Hao et~al.(2021)Hao, Lin, Post, Mikhailova, Li, Chen, Yu, and
  Liu]{hao2021}
Zhenbang Hao, Lili Lin, Christopher~J. Post, Elena~A. Mikhailova, Minghui Li,
  Yan Chen, Kunyong Yu, and Jian Liu.
\newblock Automated tree-crown and height detection in a young forest
  plantation using mask region-based convolutional neural network ({Mask
  R-CNN}).
\newblock \emph{ISPRS J. Photogramm. Remote Sens.}, 178:\penalty0 112--123,
  2021.

\bibitem[He et~al.(2013)He, Sun, and Tang]{guidedfilter}
Kaiming He, Jian Sun, and Xiaoou Tang.
\newblock Guided image filtering.
\newblock \emph{IEEE Transactions on Pattern Analysis and Machine
  Intelligence}, 35\penalty0 (6):\penalty0 1397--1409, 2013.

\bibitem[He et~al.(2020)He, Gkioxari, Doll\'{a}r, and Girshick]{maskrcnn}
Kaiming He, Georgia Gkioxari, Piotr Doll\'{a}r, and Ross Girshick.
\newblock {Mask R-CNN}.
\newblock \emph{IEEE Trans. Pattern Anal. Mach. Intell.}, 42\penalty0
  (2):\penalty0 386--397, 2020.

\bibitem[Hiernaux et~al.(2023)Hiernaux, Issoufou, Igel, Kariryaa, Kourouma,
  Chave, Mougin, and Savadogo]{hiernaux2023}
Pierre Hiernaux, Hassane Bil-Assanou Issoufou, Christian Igel, Ankit Kariryaa,
  Moussa Kourouma, J\'{e}r\^{o}me Chave, Eric Mougin, and Patrice Savadogo.
\newblock Allometric equations to estimate the dry mass of {Sahel} woody plants
  mapped with very-high resolution satellite imagery.
\newblock \emph{For. Ecol. Manage.}, 529:\penalty0 120653, 2023.

\bibitem[Huang et~al.(2026)Huang, Ding, Xiao, Liu, and Sun]{treesam}
Chen Huang, Ying Ding, Kun Xiao, Rong Liu, and Ying Sun.
\newblock Ladder-side-tuning of visual foundation model for city-scale
  individual tree detection from high-resolution remote sensing images.
\newblock \emph{Remote Sens.}, 18\penalty0 (5):\penalty0 819, 2026.

\bibitem[Huang et~al.(2016)Huang, Yu, Guan, Wang, and Guo]{huang2016}
Jianping Huang, Haipeng Yu, Xiaodan Guan, Guoyin Wang, and Ruixia Guo.
\newblock Accelerated dryland expansion under climate change.
\newblock \emph{Nature Climate Change}, 6\penalty0 (2):\penalty0 166--171,
  2016.

\bibitem[{Humanitarian OpenStreetMap Team}(2026)]{oam}
{Humanitarian OpenStreetMap Team}.
\newblock {OpenAerialMap}.
\newblock \url{https://openaerialmap.org/}, 2026.
\newblock Accessed on 26 September 2026.

\bibitem[Jackson et~al.(2005)Jackson, Jobb{\'a}gy, Avissar, Roy, Barrett, Cook,
  Farley, le~Maitre, McCarl, and Murray]{jackson2005}
Robert~B. Jackson, Esteban~G. Jobb{\'a}gy, Roni Avissar, Somnath~Baidya Roy,
  Damian~J. Barrett, Charles~W. Cook, Kathleen~A. Farley, David~C. le Maitre,
  Bruce~A. McCarl, and Brian~C. Murray.
\newblock Trading water for carbon with biological carbon sequestration.
\newblock \emph{Science}, 310\penalty0 (5756):\penalty0 1944--1947, 2005.

\bibitem[Jain and Grauman(2013)]{jain2013}
Suyog~Dutt Jain and Kristen Grauman.
\newblock Predicting sufficient annotation strength for interactive foreground
  segmentation.
\newblock In \emph{IEEE International Conference on Computer Vision (ICCV)},
  pages 1313--1320, 2013.

\bibitem[Jenkins et~al.(2013)Jenkins, Pimm, and Joppa]{jenkins2013}
Clinton~N. Jenkins, Stuart~L. Pimm, and Lucas~N. Joppa.
\newblock Global patterns of terrestrial vertebrate diversity and conservation.
\newblock \emph{Proc. Natl. Acad. Sci. U.S.A.}, 110\penalty0 (28):\penalty0
  E2602--E2610, 2013.

\bibitem[Khoreva et~al.(2017)Khoreva, Benenson, Hosang, Hein, and
  Schiele]{khoreva2017}
Anna Khoreva, Rodrigo Benenson, Jan Hosang, Matthias Hein, and Bernt Schiele.
\newblock Simple does it: Weakly supervised instance and semantic segmentation.
\newblock In \emph{IEEE Conference on Computer Vision and Pattern Recognition
  (CVPR)}, pages 876--885, 2017.

\bibitem[Kirillov et~al.(2023)Kirillov, Mintun, Ravi, Mao, Rolland, Gustafson,
  Xiao, Whitehead, Berg, Lo, Doll\'{a}r, and Girshick]{sam}
Alexander Kirillov, Eric Mintun, Nikhila Ravi, Hanzi Mao, Chloe Rolland, Laura
  Gustafson, Tete Xiao, Spencer Whitehead, Alexander~C. Berg, Wan-Yen Lo, Piotr
  Doll\'{a}r, and Ross Girshick.
\newblock Segment anything.
\newblock In \emph{IEEE/CVF International Conference on Computer Vision
  (ICCV)}, pages 3992--4003, 2023.

\bibitem[Krizhevsky et~al.(2012)Krizhevsky, Sutskever, and Hinton]{alexnet}
Alex Krizhevsky, Ilya Sutskever, and Geoffrey~E. Hinton.
\newblock {ImageNet} classification with deep convolutional neural networks.
\newblock In \emph{Advances in Neural Information Processing Systems 25 (NIPS
  2012)}, pages 1097--1105, 2012.

\bibitem[Laestadius et~al.(2012)Laestadius, Maginnis, Minnemeyer, Potapov,
  Saint-Laurent, and Sizer]{laestadius2012}
Lars Laestadius, Stewart Maginnis, Susan Minnemeyer, Peter Potapov, Carole
  Saint-Laurent, and Nigel Sizer.
\newblock Mapping opportunities for forest landscape restoration.
\newblock \emph{Unasylva}, 62\penalty0 (238):\penalty0 47--48, 2012.

\bibitem[Lai et~al.(2025)Lai, Wu, Zhan, Li, Gao, Liu, Zhang, and Guo]{boxseg}
Jinxiang Lai, Wenlong Wu, Jiawei Zhan, Jian Li, Bin-Bin Gao, Jun Liu, Jie
  Zhang, and Song Guo.
\newblock {BoxSeg}: Quality-aware and peer-assisted learning for box-supervised
  instance segmentation.
\newblock \emph{arXiv}, 2025.
\newblock arXiv:2504.05137.

\bibitem[Lan et~al.(2021)Lan, Yu, Choy, Radhakrishnan, Liu, Zhu, Davis, and
  Anandkumar]{discobox}
Shiyi Lan, Zhiding Yu, Christopher Choy, Subhashree Radhakrishnan, Guilin Liu,
  Yuke Zhu, Larry~S. Davis, and Anima Anandkumar.
\newblock {DiscoBox}: Weakly supervised instance segmentation and semantic
  correspondence from box supervision.
\newblock In \emph{IEEE/CVF International Conference on Computer Vision
  (ICCV)}, pages 3386--3396, 2021.

\bibitem[LeCun et~al.(1998)LeCun, Bottou, Bengio, and Haffner]{lecun1998}
Yann LeCun, L\'{e}on Bottou, Yoshua Bengio, and Patrick Haffner.
\newblock Gradient-based learning applied to document recognition.
\newblock \emph{Proc. IEEE}, 86\penalty0 (11):\penalty0 2278--2324, 1998.

\bibitem[Li et~al.(2024)Li, Liu, Zhu, Cui, Yu, Hua, and Zhang]{box2mask}
Wentong Li, Wenyu Liu, Jianke Zhu, Miaomiao Cui, Risheng Yu, Xian-Sheng Hua,
  and Lei Zhang.
\newblock {Box2Mask}: Box-supervised instance segmentation via level-set
  evolution.
\newblock \emph{IEEE Trans. Pattern Anal. Mach. Intell.}, 46\penalty0
  (7):\penalty0 5157--5173, 2024.

\bibitem[Li et~al.(2022)Li, Chai, Wang, Lei, and Zhang]{acercnn}
Yingbo Li, Guoqi Chai, Yueting Wang, Lingting Lei, and Xiaoli Zhang.
\newblock {ACE R-CNN}: An attention complementary and edge detection-based
  instance segmentation algorithm for individual tree species identification
  using {UAV} {RGB} images and {LiDAR} data.
\newblock \emph{Remote Sens.}, 14\penalty0 (13):\penalty0 3035, 2022.

\bibitem[Lin et~al.(2014)Lin, Maire, Belongie, Hays, Perona, Ramanan,
  Doll\'{a}r, and Zitnick]{coco}
Tsung-Yi Lin, Michael Maire, Serge Belongie, James Hays, Pietro Perona, Deva
  Ramanan, Piotr Doll\'{a}r, and C.~Lawrence Zitnick.
\newblock Microsoft {COCO}: Common objects in context.
\newblock In \emph{Computer Vision -- ECCV 2014}, pages 740--755, 2014.

\bibitem[Liu et~al.(2021)Liu, Lin, Cao, Hu, Wei, Zhang, Lin, and Guo]{swin}
Ze Liu, Yutong Lin, Yue Cao, Han Hu, Yixuan Wei, Zheng Zhang, Stephen Lin, and
  Baining Guo.
\newblock Swin transformer: Hierarchical vision transformer using shifted
  windows.
\newblock In \emph{IEEE/CVF International Conference on Computer Vision
  (ICCV)}, pages 10012--10022, 2021.

\bibitem[Lungo~Vaschetti et~al.(2025)Lungo~Vaschetti, Arnaudo, and
  Rossi]{treepseco}
Jacopo Lungo~Vaschetti, Edoardo Arnaudo, and Claudio Rossi.
\newblock {TreePseCo}: Scaling individual tree crown segmentation using large
  vision models.
\newblock \emph{Int. Arch. Photogramm. Remote Sens. Spatial Inf. Sci.},
  XLVIII-M-7-2025:\penalty0 275--282, 2025.

\bibitem[Melas-Kyriazi et~al.(2022)Melas-Kyriazi, Rupprecht, Laina, and
  Vedaldi]{deepspectral}
Luke Melas-Kyriazi, Christian Rupprecht, Iro Laina, and Andrea Vedaldi.
\newblock Deep spectral methods: A surprisingly strong baseline for
  unsupervised semantic segmentation and localization.
\newblock In \emph{IEEE/CVF Conference on Computer Vision and Pattern
  Recognition (CVPR)}, 2022.
\newblock arXiv:2205.07839.

\bibitem[Mugabowindekwe et~al.(2023)Mugabowindekwe, Brandt, Chave, Reiner,
  Skole, Kariryaa, Igel, Hiernaux, Ciais, Mertz, Tong, Li, Rwanyiziri,
  Dushimiyimana, Ndoli, Uwizeyimana, Lilles\o, Gieseke, Tucker, Saatchi, and
  Fensholt]{mugabowindekwe2023}
Maurice Mugabowindekwe, Martin Brandt, J\'{e}r\^{o}me Chave, Florian Reiner,
  David~L. Skole, Ankit Kariryaa, Christian Igel, Pierre Hiernaux, Philippe
  Ciais, Ole Mertz, Xiaoye Tong, Sizhuo Li, Gaspard Rwanyiziri, Thaulin
  Dushimiyimana, Alain Ndoli, Valens Uwizeyimana, Jens-Peter~Barnekow Lilles\o,
  Fabian Gieseke, Compton~J. Tucker, Sassan Saatchi, and Rasmus Fensholt.
\newblock Nation-wide mapping of tree-level aboveground carbon stocks in
  {Rwanda}.
\newblock \emph{Nat. Clim. Change}, 13\penalty0 (1):\penalty0 91--97, 2023.

\bibitem[Ocer et~al.(2020)Ocer, Kaplan, Erdem, Kucuk~Matci, and
  Avdan]{ocer2020}
Nuri~Erkin Ocer, Gordana Kaplan, Firat Erdem, Dilek Kucuk~Matci, and Ugur
  Avdan.
\newblock Tree extraction from multi-scale {UAV} images using {Mask R-CNN} with
  {FPN}.
\newblock \emph{Remote Sens. Lett.}, 11\penalty0 (9):\penalty0 847--856, 2020.

\bibitem[Olofsson et~al.(2014)Olofsson, Foody, Herold, Stehman, Woodcock, and
  Wulder]{olofsson2014}
Pontus Olofsson, Giles~M. Foody, Martin Herold, Stephen~V. Stehman, Curtis~E.
  Woodcock, and Michael~A. Wulder.
\newblock Good practices for estimating area and assessing accuracy of land
  change.
\newblock \emph{Remote Sensing of Environment}, 148:\penalty0 42--57, 2014.

\bibitem[Olson et~al.(2001)Olson, Dinerstein, Wikramanayake, Burgess, Powell,
  Underwood, D'Amico, Itoua, Strand, Morrison, Loucks, Allnutt, Ricketts, Kura,
  Lamoreux, Wettengel, Hedao, and Kassem]{olson2001}
David~M. Olson, Eric Dinerstein, Eric~D. Wikramanayake, Neil~D. Burgess, George
  V.~N. Powell, Emma~C. Underwood, Jennifer~A. D'Amico, Illanga Itoua, Holly~E.
  Strand, John~C. Morrison, Colby~J. Loucks, Thomas~F. Allnutt, Taylor~H.
  Ricketts, Yumiko Kura, John~F. Lamoreux, Wesley~W. Wettengel, Prashant Hedao,
  and Kenneth~R. Kassem.
\newblock Terrestrial ecoregions of the world: A new map of life on {Earth}.
\newblock \emph{BioScience}, 51\penalty0 (11):\penalty0 933--938, 2001.

\bibitem[Oquab et~al.(2024)Oquab, Darcet, Moutakanni, Vo, Szafraniec, Khalidov,
  Fernandez, Haziza, Massa, El-Nouby, et~al.]{dinov2}
Maxime Oquab, Timoth\'{e}e Darcet, Th\'{e}o Moutakanni, Huy~V. Vo, Marc
  Szafraniec, Vasil Khalidov, Pierre Fernandez, Daniel Haziza, Francisco Massa,
  Alaaeldin El-Nouby, et~al.
\newblock {DINOv2}: Learning robust visual features without supervision.
\newblock \emph{Trans. Mach. Learn. Res.}, 2024.
\newblock arXiv:2304.07193.

\bibitem[Papandreou et~al.(2015)Papandreou, Chen, Murphy, and Yuille]{emadapt}
George Papandreou, Liang-Chieh Chen, Kevin~P. Murphy, and Alan~L. Yuille.
\newblock Weakly- and semi-supervised learning of a deep convolutional network
  for semantic image segmentation.
\newblock In \emph{IEEE International Conference on Computer Vision (ICCV)},
  pages 1742--1750, 2015.

\bibitem[Pesonen et~al.(2026)Pesonen, Rua, Taher, Koivum\"{a}ki, Yu, Hakala,
  and Honkavaara]{pesonen2026}
Julius Pesonen, Stefan Rua, Josef Taher, Niko Koivum\"{a}ki, Xiaowei Yu, Teemu
  Hakala, and Eija Honkavaara.
\newblock Learning image-based tree crown segmentation from enhanced
  lidar-based pseudo-labels.
\newblock \emph{arXiv}, 2026.
\newblock arXiv:2602.13022.

\bibitem[Ple\c{s}oianu et~al.(2020)Ple\c{s}oianu, Stupariu, \c{S}andric,
  P\u{a}tru-Stupariu, and Dr\u{a}gu\c{t}]{plesoianu2020}
Alin-Ionu\c{t} Ple\c{s}oianu, Mihai-Sorin Stupariu, Ionu\c{t} \c{S}andric,
  Ileana P\u{a}tru-Stupariu, and Lucian Dr\u{a}gu\c{t}.
\newblock Individual tree-crown detection and species classification in very
  high-resolution remote sensing imagery using a deep learning ensemble model.
\newblock \emph{Remote Sens.}, 12\penalty0 (15):\penalty0 2426, 2020.

\bibitem[Pont-Tuset et~al.(2017)Pont-Tuset, Arbel{\'a}ez, Barron, Marques, and
  Malik]{pont2017mcg}
Jordi Pont-Tuset, Pablo Arbel{\'a}ez, Jonathan~T. Barron, Ferran Marques, and
  Jitendra Malik.
\newblock Multiscale combinatorial grouping for image segmentation and object
  proposal generation.
\newblock \emph{IEEE Transactions on Pattern Analysis and Machine
  Intelligence}, 39\penalty0 (1):\penalty0 128--140, 2017.

\bibitem[Que et~al.(2026)Que, Gao, Shan, Liu, An, Deng, Feng, Yang, and
  Mu]{fmsam}
Haohua Que, Haojia Gao, Weihao Shan, Mingkai Liu, Jianshuo An, Fei Deng, Shuo
  Feng, Xinghua Yang, and Lei Mu.
\newblock {FM-SAM}: Individual tree crown delineation and classification based
  on segmentation anything model ({SAM}) and {YOLOv10} in {UAV} imagery for
  forest monitoring.
\newblock \emph{Comput. Electron. Agric.}, 240:\penalty0 111162, 2026.

\bibitem[Ren et~al.(2017)Ren, He, Girshick, and Sun]{fasterrcnn}
Shaoqing Ren, Kaiming He, Ross Girshick, and Jian Sun.
\newblock {Faster R-CNN}: Towards real-time object detection with region
  proposal networks.
\newblock \emph{IEEE Trans. Pattern Anal. Mach. Intell.}, 39\penalty0
  (6):\penalty0 1137--1149, 2017.

\bibitem[Ren et~al.(2024)Ren, Luzi, Lahrichi, Kassaw, Collins, Bradbury, and
  Malof]{ren2024space}
Simiao Ren, Francesco Luzi, Saad Lahrichi, Kaleb Kassaw, Leslie~M. Collins,
  Kyle Bradbury, and Jordan~M. Malof.
\newblock Segment anything, from space?
\newblock In \emph{IEEE/CVF Winter Conference on Applications of Computer
  Vision (WACV)}, pages 8340--8350, 2024.

\bibitem[Rother et~al.(2004)Rother, Kolmogorov, and Blake]{rother2004}
Carsten Rother, Vladimir Kolmogorov, and Andrew Blake.
\newblock {GrabCut}: Interactive foreground extraction using iterated graph
  cuts.
\newblock \emph{ACM Transactions on Graphics}, 23\penalty0 (3):\penalty0
  309--314, 2004.

\bibitem[Shi et~al.(2026)Shi, Shi, Su, Li, Liu, Wan, and Zhou]{crownvim}
Erkang Shi, Ziyang Shi, Fulin Su, Lin Li, Ruifeng Liu, Fangying Wan, and Kai
  Zhou.
\newblock {CrownViM}: Context clustering meets vision {Mamba} for precise tree
  crown segmentation in aerial {RGB} imagery.
\newblock \emph{Remote Sens.}, 18\penalty0 (6):\penalty0 860, 2026.

\bibitem[Sim{\'e}oni et~al.(2021)Sim{\'e}oni, Puy, Vo, Roburin, Gidaris,
  Bursuc, P{\'e}rez, Marlet, and Ponce]{lost}
Oriane Sim{\'e}oni, Gilles Puy, Huy~V. Vo, Simon Roburin, Spyros Gidaris,
  Andrei Bursuc, Patrick P{\'e}rez, Renaud Marlet, and Jean Ponce.
\newblock Localizing objects with self-supervised transformers and no labels.
\newblock In \emph{British Machine Vision Conference (BMVC)}, 2021.
\newblock arXiv:2109.14279.

\bibitem[Sim\'{e}oni et~al.(2025)Sim\'{e}oni, Vo, Seitzer, Baldassarre, Oquab,
  Jose, Khalidov, Szafraniec, Yi, Ramamonjisoa, et~al.]{dinov3}
Oriane Sim\'{e}oni, Huy~V. Vo, Maximilian Seitzer, Federico Baldassarre, Maxime
  Oquab, Cijo Jose, Vasil Khalidov, Marc Szafraniec, Seungeun Yi, Micha\"{e}l
  Ramamonjisoa, et~al.
\newblock {DINOv3}.
\newblock \emph{arXiv}, 2025.
\newblock arXiv:2508.10104.

\bibitem[Skole et~al.(2021)Skole, Samek, Dieng, and Mbow]{skole2021}
David~L. Skole, Jay~H. Samek, Moussa Dieng, and Cheikh Mbow.
\newblock The contribution of trees outside of forests to landscape carbon and
  climate change mitigation in {West Africa}.
\newblock \emph{Forests}, 12\penalty0 (12):\penalty0 1652, 2021.

\bibitem[Soomro et~al.(2026)Soomro, Clarke, Medway, Liang, Summerhayes,
  Guerschman, de~Ligt, Armitage, and Ayers]{soomro2026}
Toufique~A. Soomro, Allister Clarke, Jonathan Medway, Bin Liang, Stephen
  Summerhayes, Juan~Pablo Guerschman, Robert de Ligt, Hugh Armitage, and
  Clinton Ayers.
\newblock {UAV}-based remote sensing for rangeland monitoring, a generalized
  and transparent workflow with an {Australian} lead.
\newblock \emph{Ecological Informatics}, 94:\penalty0 103663, 2026.

\bibitem[S{\"o}rensen(2007)]{sorensen2007}
Levke S{\"o}rensen.
\newblock A spatial analysis approach to the global delineation of dryland
  areas of relevance to the {CBD} programme of work on dry and subhumid lands.
\newblock Technical report, UNEP World Conservation Monitoring Centre,
  Cambridge, UK, 2007.
\newblock \url{https://resources.unep-wcmc.org/products/WCMC_RT111}.

\bibitem[Steier et~al.(2024)Steier, Goebel, and Iwaszczuk]{steier2024}
Janik Steier, Mona Goebel, and Dorota Iwaszczuk.
\newblock Is your training data really ground truth? a quality assessment of
  manual annotation for individual tree crown delineation.
\newblock \emph{Remote Sens.}, 16\penalty0 (15):\penalty0 2786, 2024.

\bibitem[Teng et~al.(2025)Teng, Ouaknine, Lalibert\'{e}, Bengio, Rolnick, and
  Larochelle]{teng2025}
M\'{e}lisande Teng, Arthur Ouaknine, Etienne Lalibert\'{e}, Yoshua Bengio,
  David Rolnick, and Hugo Larochelle.
\newblock Bringing {SAM} to new heights: Leveraging elevation data for tree
  crown segmentation from drone imagery.
\newblock In \emph{Advances in Neural Information Processing Systems 38
  (NeurIPS 2025)}, pages 9481--9513, 2025.

\bibitem[Tian et~al.(2021)Tian, Shen, Wang, and Chen]{boxinst}
Zhi Tian, Chunhua Shen, Xinlong Wang, and Hao Chen.
\newblock {BoxInst}: High-performance instance segmentation with box
  annotations.
\newblock In \emph{IEEE/CVF Conference on Computer Vision and Pattern
  Recognition (CVPR)}, pages 5439--5448, 2021.

\bibitem[Tong and Zhang(2025)]{tong2025}
Fei Tong and Yun Zhang.
\newblock Individual tree crown delineation in high resolution aerial {RGB}
  imagery using {StarDist}-based model.
\newblock \emph{Remote Sens. Environ.}, 319:\penalty0 114618, 2025.

\bibitem[Tucker et~al.(2023)Tucker, Brandt, Hiernaux, Kariryaa, Rasmussen,
  Small, Igel, Reiner, Melocik, Meyer, Sinno, Romero, Glennie, Fitts, Morin,
  Pinzon, McClain, Morin, Porter, Loeffler, Kergoat, Issoufou, Savadogo,
  Wigneron, Poulter, Ciais, Kaufmann, Myneni, Saatchi, and
  Fensholt]{tucker2023}
Compton Tucker, Martin Brandt, Pierre Hiernaux, Ankit Kariryaa, Kjeld
  Rasmussen, Jennifer Small, Christian Igel, Florian Reiner, Katherine Melocik,
  Jesse Meyer, Scott Sinno, Eric Romero, Erin Glennie, Yasmin Fitts, August
  Morin, Jorge Pinzon, Devin McClain, Paul Morin, Claire Porter, Shane
  Loeffler, Laurent Kergoat, Bil-Assanou Issoufou, Patrice Savadogo,
  Jean-Pierre Wigneron, Benjamin Poulter, Philippe Ciais, Robert Kaufmann,
  Ranga Myneni, Sassan Saatchi, and Rasmus Fensholt.
\newblock Sub-continental-scale carbon stocks of individual trees in {African}
  drylands.
\newblock \emph{Nature}, 615\penalty0 (7950):\penalty0 80--86, 2023.

\bibitem[Veitch-Michaelis et~al.(2024)Veitch-Michaelis, Cottam, Schweizer,
  Broadbent, Dao, Zhang, Almeyda~Zambrano, and Max]{oamtcd}
Josh Veitch-Michaelis, Andrew Cottam, Daniella Schweizer, Eben~N. Broadbent,
  David Dao, Ce Zhang, Angelica Almeyda~Zambrano, and Simeon Max.
\newblock {OAM-TCD}: A globally diverse dataset of high-resolution tree cover
  maps.
\newblock In \emph{Advances in Neural Information Processing Systems 37
  (NeurIPS 2024), Datasets and Benchmarks Track}, pages 49749--49767, 2024.

\bibitem[Veldman et~al.(2015)Veldman, Overbeck, Negreiros, Mahy, Le~Stradic,
  Fernandes, Durigan, Buisson, Putz, and Bond]{veldman2015}
Joseph~W. Veldman, Gerhard~E. Overbeck, Daniel Negreiros, Gregory Mahy, Soizig
  Le~Stradic, G.~Wilson Fernandes, Giselda Durigan, Elise Buisson, Francis~E.
  Putz, and William~J. Bond.
\newblock Where tree planting and forest expansion are bad for biodiversity and
  ecosystem services.
\newblock \emph{BioScience}, 65\penalty0 (10):\penalty0 1011--1018, 2015.

\bibitem[Wang et~al.(2023)Wang, Girdhar, Yu, and Misra]{cutler}
Xudong Wang, Rohit Girdhar, Stella~X. Yu, and Ishan Misra.
\newblock Cut and learn for unsupervised object detection and instance
  segmentation.
\newblock In \emph{IEEE/CVF Conference on Computer Vision and Pattern
  Recognition (CVPR)}, 2023.
\newblock arXiv:2301.11320.

\bibitem[Wang et~al.(2022)Wang, Shen, Hu, Yuan, Crowley, and
  Vaufreydaz]{tokencut}
Yangtao Wang, Xi Shen, Shell~Xu Hu, Yuan Yuan, James~L. Crowley, and Dominique
  Vaufreydaz.
\newblock Self-supervised transformers for unsupervised object discovery using
  normalized cut.
\newblock In \emph{IEEE/CVF Conference on Computer Vision and Pattern
  Recognition (CVPR)}, 2022.
\newblock arXiv:2202.11539.

\bibitem[Wei et~al.(2017)Wei, Zhang, Li, Xie, Wu, Shen, and Zhou]{ddt}
Xiu-Shen Wei, Chen-Lin Zhang, Yao Li, Chen-Wei Xie, Jianxin Wu, Chunhua Shen,
  and Zhi-Hua Zhou.
\newblock Deep descriptor transforming for image co-localization.
\newblock In \emph{26th International Joint Conference on Artificial
  Intelligence (IJCAI)}, pages 3048--3054, 2017.

\bibitem[Weinstein et~al.(2020)Weinstein, Marconi, Aubry-Kientz, Vincent,
  Senyondo, and White]{weinstein2020}
Ben~G. Weinstein, Sergio Marconi, M\'{e}laine Aubry-Kientz, Gregoire Vincent,
  Henry Senyondo, and Ethan~P. White.
\newblock {DeepForest}: A {Python} package for {RGB} deep learning tree crown
  delineation.
\newblock \emph{Methods Ecol. Evol.}, 11\penalty0 (12):\penalty0 1743--1751,
  2020.

\bibitem[Weyler et~al.(2024)Weyler, Magistri, Marks, Chong, Sodano, Roggiolani,
  Chebrolu, Stachniss, and Behley]{phenobench}
Jan Weyler, Federico Magistri, Elias Marks, Yue~Linn Chong, Matteo Sodano,
  Gianmarco Roggiolani, Nived Chebrolu, Cyrill Stachniss, and Jens Behley.
\newblock {PhenoBench}: A large dataset and benchmarks for semantic image
  interpretation in the agricultural domain.
\newblock \emph{IEEE Transactions on Pattern Analysis and Machine
  Intelligence}, 46\penalty0 (12):\penalty0 9583--9594, 2024.

\bibitem[Wilaiwongsakul et~al.(2026)Wilaiwongsakul, Liang, Jia, Zheng, and
  Chen]{bare}
Attavit Wilaiwongsakul, Bin Liang, Wenfeng Jia, Bryan Zheng, and Fang Chen.
\newblock {BARE}: Boundary-aware with resolution enhancement for tree crown
  delineation.
\newblock In \emph{Data Science and Machine Learning (AusDM 2025)}, pages
  302--315, Singapore, 2026. Springer Nature Singapore.

\bibitem[Woebbecke et~al.(1995)Woebbecke, Meyer, Von~Bargen, and
  Mortensen]{woebbecke1995}
D.~M. Woebbecke, G.~E. Meyer, K. Von~Bargen, and D.~A. Mortensen.
\newblock Color indices for weed identification under various soil, residue,
  and lighting conditions.
\newblock \emph{Transactions of the ASAE}, 38\penalty0 (1):\penalty0 259--269,
  1995.

\bibitem[Wu et~al.(2019)Wu, Kirillov, Massa, Lo, and Girshick]{detectron2}
Yuxin Wu, Alexander Kirillov, Francisco Massa, Wan-Yen Lo, and Ross Girshick.
\newblock {Detectron2}.
\newblock \url{https://github.com/facebookresearch/detectron2}, 2019.
\newblock Accessed on 4 September 2026.

\bibitem[Yang et~al.(2023)Yang, Song, Ge, and Li]{boxsnake}
Rui Yang, Lin Song, Yixiao Ge, and Xiu Li.
\newblock {BoxSnake}: Polygonal instance segmentation with box supervision.
\newblock In \emph{IEEE/CVF International Conference on Computer Vision
  (ICCV)}, pages 766--776, 2023.

\bibitem[Zhang et~al.(2023)Zhang, Li, Liu, Zhang, Su, Zhu, Ni, and
  Shum]{dinodetr}
Hao Zhang, Feng Li, Shilong Liu, Lei Zhang, Hang Su, Jun Zhu, Lionel~M. Ni, and
  Heung-Yeung Shum.
\newblock {DINO}: {DETR} with improved denoising anchor boxes for end-to-end
  object detection.
\newblock In \emph{International Conference on Learning Representations
  (ICLR)}, 2023.
\newblock arXiv:2203.03605.

\bibitem[Zhang et~al.(2024)Zhang, Shao, Wan, Meng, Cao, and
  Wang]{zhang2024boundary}
Jie Zhang, Mingwen Shao, Yecong Wan, Lingzhuang Meng, Xiangyong Cao, and
  Shuigen Wang.
\newblock Boundary-aware spatial and frequency dual-domain transformer for
  remote sensing urban images segmentation.
\newblock \emph{IEEE Trans. Geosci. Remote Sens.}, 62:\penalty0 1--18, 2024.

\bibitem[Zhang et~al.(2026)Zhang, Zhong, Fan, Tjahjadi, Fu, and
  Ye]{treeadapter}
Jiuyu Zhang, Xingpei Zhong, Xijian Fan, Tardi Tjahjadi, Liyong Fu, and Qiaolin
  Ye.
\newblock Adapting pre-trained large model to robust individual tree crown
  detection from {UAV} optical images.
\newblock \emph{Smart Agric. Technol.}, 13:\penalty0 101674, 2026.

\bibitem[Zhao et~al.(2023)Zhao, Morgenroth, Pearse, and Schindler]{zhao2023}
Haotian Zhao, Justin Morgenroth, Grant Pearse, and Jan Schindler.
\newblock A systematic review of individual tree crown detection and
  delineation with convolutional neural networks ({CNN}).
\newblock \emph{Curr. For. Rep.}, 9\penalty0 (3):\penalty0 149--170, 2023.

\bibitem[Zhu et~al.(2017)Zhu, Tuia, Mou, Xia, Zhang, Xu, and
  Fraundorfer]{zhu2017}
Xiao~Xiang Zhu, Devis Tuia, Lichao Mou, Gui-Song Xia, Liangpei Zhang, Feng Xu,
  and Friedrich Fraundorfer.
\newblock Deep learning in remote sensing: A comprehensive review and list of
  resources.
\newblock \emph{IEEE Geosci. Remote Sens. Mag.}, 5\penalty0 (4):\penalty0
  8--36, 2017.

\end{thebibliography}
}

\clearpage
\appendix
\renewcommand{\thetable}{A\arabic{table}}
\renewcommand{\thefigure}{A\arabic{figure}}
\renewcommand{\thealgorithm}{A\arabic{algorithm}}
\setcounter{table}{0}\setcounter{figure}{0}\setcounter{algorithm}{0}
\section{Additional Method and Experimental Details}
\label{app:details}

\subsection{Tree Crown Methods by Annotation Type}
\label{app:gap}

\begin{table*}[!ht]
\centering\footnotesize
\setlength{\tabcolsep}{3pt}
\renewcommand{\arraystretch}{1.12}
\caption{Tree crown methods for UAV and airborne imagery, in order of publication. \emph{Input}: at inference. \emph{Training labels} and \emph{Training data}: as published. \emph{Evaluation}: biome groups evaluated in the original publication. Table~\ref{tab:oamtcd439} compares every tree crown method here that outputs masks, takes only RGB at inference, can be trained on OAM-TCD's box or polygon labels without LiDAR, and has runnable public code, alongside the generic box-supervised methods Box2Mask~\cite{box2mask} and BoxTeacher~\cite{boxteacher}; SAM\,3~\cite{sam3} is compared as a box-prompted masker (Table~\ref{tab:samebox}).}
\label{tab:gap}
\renewcommand{\tabularxcolumn}[1]{m{#1}}
\begin{tabularx}{\textwidth}{@{}>{\raggedright\arraybackslash}m{2.6cm} l >{\raggedright\arraybackslash}m{2.6cm} >{\raggedright\arraybackslash}X l ccc@{}}
\toprule
 & & & & & \multicolumn{3}{c}{Evaluation} \\
\cmidrule(l){6-8}
Method & Input & Training labels & Training data & Out & Tr & Te & Sv \\
\midrule
DeepForest \cite{weinstein2020} & RGB & ${>}10$k hand boxes, LiDAR-derived boxes & NEON, $22$ US forest sites & Box & \cmark & \cmark & \cmark$^{a}$ \\
SSD ensemble \cite{plesoianu2020} & RGB + DSM & boxes & orchard and forest, Romania and Germany & Box & \xmark & \cmark & \xmark \\
ACE R-CNN \cite{acercnn} & RGB + CHM & polygons & plantations, Guangxi, China & Mask & \cmark & \xmark & \xmark \\
Detectree2 \cite{ball2023} & RGB & $3{,}797$ polygons & tropical forest; tiles with ${<}40\%$ crown cover excluded & Mask & \cmark & \xmark & \xmark \\
Mask R-CNN (OAM-TCD) \cite{oamtcd} & RGB & polygons & OAM-TCD, global, all biomes & Mask & \omark & \omark & \omark \\
StarDist \cite{tong2025} & RGB & polygons & mixed forest, Canada & Mask & \xmark & \cmark & \xmark \\
TreePseCo \cite{treepseco} & RGB & LiDAR treetops, NEON boxes & NEON, US forest sites & Mask & \xmark & \cmark & \omark \\
BalSAM \cite{teng2025} & RGB + DSM & polygons & boreal, temperate and tropical forest & Mask & \cmark & \cmark & \xmark \\
CrownViM \cite{crownvim} & RGB & polygons & OAM-TCD; closed-canopy forest (SSD) & Mask & \omark & \omark & \omark \\
FG-TreeSeg \cite{fgtreeseg} & RGB & semantic canopy masks & OAM-TCD & Mask & \xmark & \cmark & \omark \\
Mask R-CNN \cite{pesonen2026} & RGB & LiDAR crowns refined by SAM\,2 & urban forest, Finland & Mask & \xmark & \cmark & \cmark$^{b}$ \\
TreeCoG \cite{treecog} & RGB & polygons & forest, Vietnam; BAMFORESTS, Germany & Mask & \cmark & \cmark & \xmark \\
Tree-SAM \cite{treesam} & RGB & polygons & urban and forest, Guangzhou, China & Mask & \cmark & \cmark & \xmark \\
TreeAdapter \cite{treeadapter} & RGB & NEON boxes & NEON, US forest sites & Box & \cmark & \cmark & \cmark \\
\midrule
\rowcolor{black!8}
\textbf{LACE (ours)} & \textbf{RGB} & \textbf{boxes} & \textbf{OAM-TCD, \boldmath$900$ images} & \textbf{Mask} & \omark & \omark & \cmark$^{c}$ \\
\bottomrule
\end{tabularx}

\vspace{3pt}
\begin{minipage}{0.95\textwidth}
\footnotesize
Biome groups of World Wildlife Fund (WWF) terrestrial biomes~\cite{olson2001} as used throughout: Tr, tropical forest (WWF $1$--$3$, $14$); Te, temperate and boreal forest ($4$--$6$, $11$); Sv, savanna/dryland ($7$--$10$, $12$, $13$). $\bullet$ reported separately; $\circ$ only within an aggregate benchmark; $\times$ not evaluated. DSM, digital surface model; CHM, canopy height model. $^{a}$Per-site NEON results; the worst site is a desert scrub site absent from training. $^{b}$Out-of-distribution test on a NEON oak savanna site, where the RGB model detects $2$ of $148$ trees. $^{c}$Sparse-canopy slice (Table~\ref{tab:sparse236}), tile-disjoint from training but not scene-disjoint. Crown mapping at $0.25$--$0.5$~m/px~\cite{brandt2020,braga2020,tucker2023,mugabowindekwe2023} is out of scope here.
\end{minipage}
\end{table*}

CrownViM~\cite{crownvim}, FG-TreeSeg~\cite{fgtreeseg}, StarDist~\cite{tong2025}, TreeCoG~\cite{treecog} and Tree-SAM~\cite{treesam} are not among our baselines because none has runnable public code: CrownViM's repository releases only the backbone module, without training or inference code, configuration files, trained weights or a licence, and we found no released code for the others.

\subsection{Method Details}
\label{app:method}

Registration of the interlaced grid of Equation~\eqref{eq:interlace} is verified independently: a synthetic blob is localised to within one cell across $200$ positions, and the cells of $\widetilde F$ belonging to the unshifted phase reproduce a plain single-pass extraction to floating-point precision, confirming its indexing.

Note that the standard von Mises--Fisher normalising constant $C_d(\kappa)$ is omitted from Equation~\eqref{eq:bgll} and from the foreground term of Equation~\eqref{eq:logratio} alike; sharing a single concentration parameter $\kappa$ between the two mixtures makes it identical on both sides, so it cancels in the difference and never has to be evaluated in $d = 128$ dimensions.

Separate priors are kept for three crown-size bands (training box-size terciles), since at a fixed feature stride the number of cells spanning a crown depends on its size; each predicted box is masked with the prior of its band. Each bin is also capped at $0.95$, so no relative position is ever treated as certainly foreground.
Pinning the mass to a constant derived from geometry rather than fitted is what keeps the prior label-free, and it means the fit decides only \emph{where} the foreground mass sits, never \emph{how much} of it there is: without the constraint $\rho$ inflates towards unity, stops discriminating position, and the $\alpha$ term of Equation~\eqref{eq:posterior} degenerates into an additive constant.

Strictly speaking, the procedure is EM-style rather than EM proper: the within-box mean-centring of Equation~\eqref{eq:meancentre}, the contrastive term of Equation~\eqref{eq:contrastive} and the projection of $\rho$ onto the mass and cap constraints each break the correspondence with a joint likelihood, so no objective is monotonically ascended and there is no convergence test: $T$ is fixed at $30$ iterations.
The constraints are applied as a single rescale followed by the cap, so the cap partly undoes the rescaling: on the deployed fit this leaves a realised mean per-bin mass of $0.690$ against the nominal $0.785$.

The four inputs of $\hat s$ (Section~\ref{sec:scorer}) are what remained after backward elimination from fourteen candidates on validation; tile-level context features were the first removed.

\subsection{Supervision and Model Selection}
\label{app:selection}

The prior weight $\alpha$, the concentration $\kappa$, the mask threshold $\tau$ and the radius and $\epsilon$ of the guided filter are selected on the $108$ validation images (Section~\ref{sec:labels}), on the boxes of an earlier detector trained on the same images, and were frozen before the DINO-DETR detector was trained; none of these requires refitting. The $4$~px lattice, the box-relative $\tau$ and ranking by the $4$~px posterior were chosen by a hyperparameter sweep of ${\sim}170$ configurations on the same validation images. The learned confidence $\hat s$ is trained on the validation detections with box-match labels and was selected by validation mask AP$_{50}$. We therefore describe the method as trained without mask supervision but also broadly calibrated for tree crown mask generation on a small held-out set of mask labels as a one-time analysis. We take these to be reasonable `tree crown' hyperparameters and do not attempt to optimise them for use on other datasets.

\begin{table*}[!ht]
\footnotesize
\renewcommand{\arraystretch}{1.0}
\setlength{\aboverulesep}{0.2ex}\setlength{\belowrulesep}{0.25ex}
\caption{Configuration. Encoder and masker-fitting hyperparameters were fixed a priori, and the inference settings were selected on validation with an earlier detector's boxes and frozen for the DINO-DETR detector (Section~\ref{sec:inference}). The detector follows DINO-DETR's released optimiser with a validation-plateau schedule. The learned confidence $\hat s$ is trained on validation detections (Section~\ref{sec:scorer}).\label{tab:config}} \newcolumntype{K}{>{\hsize=0.42\hsize\raggedright\arraybackslash}X} \newcolumntype{V}{>{\hsize=1.58\hsize\raggedright\arraybackslash}X}
\begin{tabularx}{\textwidth}{KV}
\toprule
\textbf{Component} & \textbf{Setting} \\
\midrule
Encoder & DINOv3 ViT-L/16, frozen, layer $24$ only, $1024$ dimensions \\
\cmidrule{1-2}
Feature grid & $4$ offsets $(\delta_y,\delta_x) \in \{0,8\}^2$, interlaced to $8$~px ($256\times256$ cells), used to fit the box-to-mask module. $16$ offsets $\{0,4,8,12\}^2$, interlaced to $4$~px ($512\times512$ cells), at inference \\
\cmidrule{1-2}
Detector & DINO-DETR (detrex), Swin-L (ImageNet-22k), $5$-scale deformable encoder, $900$ queries, denoising, two-stage. COCO-trained initialisation, $1$ class \\
\cmidrule{1-2}
Detector training & AdamW, lr $10^{-4}$ (backbone $\times0.1$), weight decay $10^{-4}$, gradient clip $0.1$, batch $16$. $3{,}168$ jittered $1024$~px crops per epoch (scale $0.8$--$1.0$, horizontal flip). lr $\times0.1$ on validation plateau ($\Delta = 0.005$, patience $2$, at most $2$ decays), $4$--$20$ epochs. Best checkpoint by validation box AP$_{50}$ on the $896$ of $972$ validation subtiles that hold any annotation \\
\cmidrule{1-2}
Decoding & $3\times3$ subtiles of $1024$~px at stride $512$, top $300$ each. Boxes within $10$~px of an interior subtile edge dropped, duplicates removed at IoU $>0.7$ or containment $>0.85$. Score $>0.05$, top $600$ per tile \\
\midrule
Masker -- feature space & $120$ training tiles; PCA $d = 128$ \\
\cmidrule{1-2}
Masker -- mixtures & $K_{fg} = 16$ foreground prototypes; $K_{bg} = 12$ background components; concentration $\kappa = 10$; contrastive weight $\beta = 0.5$ \\
\cmidrule{1-2}
Masker -- spatial prior & $N = 8$ bins per axis; $3$ size bands at the training terciles; mean per-bin foreground mass $\pi/4$; per-bin cap $0.95$ \\
\cmidrule{1-2}
Masker -- EM schedule & $30$ iterations, no convergence test; prototype pruning after $5$ iterations at responsibility share below $0.25/K_{fg}$, a quarter of uniform \\
\midrule
Inference & prior weight $\alpha = 0.3$; concentration $\kappa = 16$ (fitted at $10$); posterior upsampled to the $2048$~px tile by an RGB-guided filter (radius $12$~px, $\epsilon = 10^{-4}$); mask threshold $\tau = 0.33\times$ the in-box maximum of the filtered posterior \\
\cmidrule{1-2}
Ranking & $\hat s$: MLP with one hidden layer of $64$ units on $s$, $\log(\bar p\,D)$, $\log h$ and $\mathrm{IoU}_{\uparrow}$. Mean of $25$ MLPs ($5$ seeds $\times$ $5$ folds) trained on the $108$ validation tiles. Decode floor applied to $s$ \\
\bottomrule
\end{tabularx}
\end{table*}

\subsection{Datasets}
\label{app:datasets}

\begin{table*}[!ht]
\footnotesize\setlength{\tabcolsep}{3pt}
\centering
\caption{Biome composition of the $439$-tile OAM-TCD holdout test set by WWF biome label, grouped as in Table~\ref{tab:gap}. Crowns counts individual crown annotations. Crown px and canopy px are the mean per-tile fraction of labelled crown and labelled canopy pixels.}
\label{tab:test-biomes}
\small\setlength{\tabcolsep}{3pt}
\begin{tabular}{@{}clrrrr@{}}
\toprule
Code & Biome (WWF) & Tiles & Crowns & Crown px & Canopy px \\
\midrule
\multicolumn{6}{l}{\emph{Tropical forest (Tr)}} \\
$1$ & Trop./Subtrop. Moist Broadleaf Forest & $187$ & $16{,}855$ & $4.7\%$ & $32.8\%$ \\
$2$ & Trop./Subtrop. Dry Broadleaf Forest & $10$ & $175$ & $0.8\%$ & $4.4\%$ \\
$14$ & Mangroves & $1$ & $20$ & $0.5\%$ & $68.1\%$ \\
 & \emph{Group} & $198$ & $17{,}050$ & $4.4\%$ & $31.5\%$ \\
\midrule
\multicolumn{6}{l}{\emph{Temperate and boreal forest (Te)}} \\
$4$ & Temperate Broadleaf \& Mixed Forest & $137$ & $4{,}158$ & $2.4\%$ & $16.1\%$ \\
$5$ & Temperate Conifer Forest & $31$ & $1{,}832$ & $4.4\%$ & $7.8\%$ \\
$6$ & Boreal Forest/Taiga & $3$ & $44$ & $3.0\%$ & $10.9\%$ \\
 & \emph{Group} & $171$ & $6{,}034$ & $2.8\%$ & $14.5\%$ \\
\midrule
\multicolumn{6}{l}{\emph{Savanna/dryland (Sv)}} \\
$7$ & Trop./Subtrop. Grassland, Savanna \& Shrubland & $8$ & $82$ & $1.1\%$ & $11.1\%$ \\
$13$ & Desert \& Xeric Shrubland & $4$ & $97$ & $1.2\%$ & $2.3\%$ \\
$8$ & Temperate Grassland, Savanna \& Shrubland & $2$ & $87$ & $2.4\%$ & $5.1\%$ \\
$12$ & Mediterranean Forest, Woodland \& Scrub & $2$ & $111$ & $3.9\%$ & $9.8\%$ \\
 & \emph{Group} & $16$ & $377$ & $1.7\%$ & $8.0\%$ \\
\midrule
\multicolumn{6}{l}{\emph{No WWF biome match}} \\
-- & Unmatched & $54$ & $2{,}244$ & $2.8\%$ & $16.3\%$ \\
\cmidrule(lr){1-6}
 & \textbf{Total} & $\mathbf{439}$ & $\mathbf{25{,}705}$ & & \\
\bottomrule
\end{tabular}

\vspace{3pt}
\begin{minipage}{0.95\textwidth}
\footnotesize
WWF biomes $3$, $9$, $10$ and $11$ have no tiles in the holdout test set and are omitted.
\end{minipage}
\end{table*}

\begin{table*}[!ht]
\centering
\caption{Composition of the sparse-canopy slice. Tiles are selected by WWF biome label, excluding all $900$ training tiles. One eligible tile of biome $12$ was dropped at build time as unreadable, leaving $236$. Crown px and canopy px are the mean per-tile fraction of labelled crown and labelled canopy pixels.}
\label{tab:sparse-biomes}
\setlength{\tabcolsep}{4pt}
\begin{tabular}{@{}clrrr@{}}
\toprule
Code & Biome (WWF) & Tiles & Crown px & Canopy px \\
\midrule
\multicolumn{5}{l}{\emph{Sparse-canopy slice}} \\
$7$ & Trop./Subtrop. Grassland, Savanna \& Shrubland & $102$ & $2.3\%$ & $8.6\%$ \\
$12$ & Mediterranean Forest, Woodland \& Scrub & $56$ & $4.8\%$ & $7.6\%$ \\
$8$ & Temperate Grassland, Savanna \& Shrubland & $31$ & $4.8\%$ & $11.5\%$ \\
$13$ & Desert \& Xeric Shrubland & $24$ & $6.1\%$ & $15.2\%$ \\
$10$ & Montane Grassland \& Shrubland & $16$ & $1.1\%$ & $1.9\%$ \\
$9$ & Flooded Grassland \& Savanna & $7$ & $1.0\%$ & $0.8\%$ \\
\cmidrule(lr){1-5}
 & \textbf{Subtotal} & $\mathbf{236}$ &  & \\
\bottomrule
\end{tabular}

\vspace{3pt}
\begin{minipage}{0.95\textwidth}
\footnotesize
Excluded from the slice: the forest, mangrove, taiga and tundra biomes (WWF $1$--$6$, $11$ and $14$) and tiles without a WWF biome match. Labelled canopy averages $23.9$--$32.6\%$ per tile on the five broadleaf and conifer forest biomes (WWF $1$--$5$) over all OAM-TCD tiles outside the $900$ training images, against $0.8$--$15.2\%$ in the slice.
\end{minipage}
\end{table*}

\begin{table}[!ht]
\footnotesize\setlength{\tabcolsep}{3pt}
\caption{Protocol for the PhenoBench transfer test. \label{tab:phenobench-protocol}}
\begin{tabularx}{\linewidth}{lX}
\toprule
\textbf{Item} & \textbf{Setting} \\
\midrule
Data & PhenoBench v1.1.0 validation split, $772$ images of $1024 \times 1024$~px (test labels are withheld) \\
Prompts & Tight pixel extent of each GT plant instance; identical for every masker \\
Scored plants & $8{,}568$ fully visible plants ($4{,}967$ crop, $3{,}601$ weed); the $1{,}840$ partially visible plants are prompted but reported separately \\
LACE masker & The OAM-TCD model as deployed: same encoder, sixteen offsets, same fitted module and inference settings; nothing refitted \\
SAM~3 & Zero-shot, whole image, the same box prompts, masks scored as returned \\
Excess Green & $2g - r - b$ on chromatic coordinates, one Otsu threshold per image, clipped to the box; no tuning \\
Metric & Mean per-instance mask IoU; paired image-bootstrap $95\%$ intervals ($2{,}000$ resamples) \\
Not applicable & No detector exists for this domain, so no AP is reported and the confidence $\hat s$ plays no part \\
\bottomrule
\end{tabularx}
\noindent{\footnotesize{LACE masker $-$ SAM~3 is $-0.114$ {[$-0.120$, $-0.107$]} over all scored plants, and the gap is widest on large plants (box $\sqrt{\text{area}}$ above $104.0$~px): $0.650$ against $0.866$.}}
\end{table}

\begin{table*}[!ht]
\caption{PhenoBench was selected from GT statistics alone, before any model was run on it. Size is the median object extent in $16$~px encoder patches, fill is the fraction of the box covered by the GT mask, and crowding is the fraction of each box covered by other instances. \label{tab:phenobench-choice}}
\newcolumntype{C}{>{\centering\arraybackslash}X}
\small\setlength{\tabcolsep}{4pt}
\begin{tabularx}{\textwidth}{Xcccccc}
\toprule
& \multicolumn{4}{c}{\textbf{GT statistics}} & \multicolumn{2}{c}{\textbf{Mask IoU from GT boxes}} \\
\cmidrule(lr){2-5}\cmidrule(lr){6-7}
\textbf{Dataset} & \textbf{Size} & \textbf{Fill} & \textbf{Ellipse IoU} & \textbf{Crowding} & \textbf{LACE masker} & \textbf{SAM~3} \\
\midrule
OAM-TCD test (source) & $2.6$ & $0.66$ & $0.80$ & $0.03$ & -- & -- \\
PhenoBench validation & $3.8$ & $0.40$ & $0.44$ & $0.04$ & $0.605$ & $0.719$ \\
\bottomrule
\end{tabularx}
\noindent{\footnotesize{PhenoBench size, fill and crowding are from the $60$-image sample used for selection. OAM-TCD is evaluated on detector boxes (Tables~\ref{tab:oamtcd439} and~\ref{tab:samebox}), not from GT boxes.}}
\end{table*}

\subsection{Additional Ablations}
\label{app:ablations}

\begin{table*}[!ht]
\caption{Interlaced versus interpolated features for the box-to-mask module on the OAM-TCD $439$-tile test split. All arms mask the same frozen detections with the same module and are ranked by the detector score $s$. $\Delta$ rows are paired tile bootstrap differences ($2{,}000$ resamples) with $95\%$ intervals. \label{tab:grid}}
\newcolumntype{C}{>{\centering\arraybackslash}X}
\begin{tabularx}{\textwidth}{lCCC}
\toprule
\textbf{Feature grid} & \textbf{Mask AP$_{50}$} & \textbf{Mask AP$_{75}$} & \textbf{Mask AP$_{50:95}$} \\
\midrule
Native $16$~px, interpolated to $4$~px & $0.6872$ & $0.2454$ & $0.3229$ \\
Interlaced $4$-offset $8$~px, interpolated to $4$~px & $\mathbf{0.6901}$ & $0.2566$ & $0.3278$ \\
Interlaced $16$-offset, real $4$~px (LACE) & $0.6900$ & $\mathbf{0.2589}$ & $\mathbf{0.3288}$ \\
\midrule
$\Delta$ real $4$~px $-$ native & $+0.003$ {\scriptsize [$0.002$, $0.006$]} & $+0.014$ {\scriptsize [$0.010$, $0.018$]} & $+0.006$ {\scriptsize [$0.004$, $0.008$]} \\
$\Delta$ real $4$~px $-$ $4$-offset & $0.000$ {\scriptsize [$-0.001$, $0.002$]} & $+0.002$ {\scriptsize [$0.001$, $0.005$]} & $+0.001$ {\scriptsize [$0.000$, $0.002$]} \\
\bottomrule
\end{tabularx}
\noindent{\footnotesize{Interpolated and real cells agree at a mean cosine of $0.950$ (native) and $0.981$ ($4$-offset) away from the source lattice. Single detector run. The module's inference settings were selected on real $4$~px features, which favours the last row.}}
\end{table*}

Foreground/background commonality within a box rests to some extent on the assumption that the box contains mostly one crown, meaning crowded trees with overlapping crowns could be a challenge. We call a crown \emph{touching} when its GT box overlaps that of another crown, which happens to be the case for $39.3\%$ of the $25{,}705$ test crowns. Touching and isolated crowns score almost identically: mean mask IoU $0.7353$ (touching) against $0.7330$ (isolated), and touching crowns fail slightly more often ($0.0241$ versus $0.0223$, where failure is a box-matched true positive whose mask IoU falls below $0.5$). Touching crowns are also larger (mean box side $\sqrt{wh}$ of $68.9$ versus $42.0$~px), which may partly offset neighbour interference.

The module was designed as a mixture of $K_{fg}$ foreground prototypes, on the expectation that different tree types (e.g.\ conifers and palms) would occupy distinct prototypes. Instead, the repulsion (Equation~\eqref{eq:contrastive}) collapses the mixture onto a single `treeness' direction. Every prototype is pushed away from the outside-box features it most resembles, so as $\beta$ rises they converge (Table~\ref{tab:collapse}). At $\beta = 0.5$ the mean pairwise cosine is $0.994$ and the effective rank $1.66$ of $16$, and refits at three EM seeds and two values of $K_{fg}$ recover the same direction ($\cos \geq 0.9908$). At $\beta = 0$ the same foreground is spread over sixteen prototypes with no fixed identities. Pruning removes only starved prototypes, so it acts at intermediate $\beta$ but not on the near-identical prototypes at $\beta = 0.5$, which share responsibility equally.

\begin{table*}[!ht]
\caption{Foreground prototype geometry and accuracy, E-step mean-centring held on throughout so that $\beta$ is the only variable. Accuracy is mean per-crown mask IoU on the GT boxes of the $108$ validation tiles at EM seed $0$, matching the geometry columns. Effective rank is $\exp$ of the entropy of the normalised singular-value spectrum of the prototype matrix i.e. it counts directions.
\label{tab:collapse}}
\newcolumntype{C}{>{\centering\arraybackslash}X}
\begin{tabularx}{\textwidth}{lCCCCC}
\toprule
\textbf{Config.} & \textbf{$K_{fg}$ init} & \textbf{$K_{fg}$ final} & \textbf{Eff. rank} & \textbf{Pairwise $\cos$} & \textbf{Crown IoU} \\
\midrule
$\beta = 0$    & $16$ & $16$ & $11.43$ & $0.234$ & $0.7950$ \\
$\beta = 0.1$  & $16$ & $15$ & $9.41$  & $0.334$ & $0.7953$ \\
$\beta = 0.25$ & $16$ & $13$ & $6.37$  & $0.525$ & $0.7958$ \\
$\beta = 0.5$  & $16$ & $16$ & $\mathbf{1.66}$ & $\mathbf{0.994}$ & $0.7980$ \\
\midrule
$\beta = 0$    & $2$  & $2$  & $2.00$  & $0.053$ & $0.7980$ \\
$\beta = 0.5$  & $2$  & $2$  & $1.30$  & $0.988$ & $0.7985$ \\
\bottomrule
\end{tabularx}
\noindent{\footnotesize{Effective rank falls sevenfold across the $K_{fg}=16$ block while crown IoU moves by at most $0.003$. At $\beta = 0.5$ both $K_{fg}$ settings collapse, and the two reach the same direction ($\cos(\bar{C}) = 0.9942$, exceeding the between-seed agreement within $K_{fg}=16$ itself, min $0.9915$, mean $0.9922$ over three EM seeds).}}
\end{table*}

Accuracy is flat throughout. Crown IoU spans $0.0035$ across all six fits, within EM-seed variation, and $K_{fg} = 2$ matches $K_{fg} = 16$ at both values of $\beta$. Mean-centring, on in every row, does most of the work (Section~\ref{sec:ablation}). We kept $K_{fg} = 16$ as the design prior and $\beta = 0.5$ from development, but with mean-centring on, either value of $\beta$ or $K_{fg}$ can be used.

\section{Scoring Protocol and Baseline Configurations}
\label{app:protocol}

\subsection{Frozen Scoring Protocol}

Every OAM-TCD figure in this paper is produced by one scorer under one configuration, fixed before the baselines were run and unchanged since, apart from a floating-point correction to the IoU thresholds worth $0.0005$ AP$_{50:95}$. The evaluation engine is unmodified \texttt{pycocotools} \texttt{COCOeval}; what follows is its configuration (Table~\ref{tab:protocol}) and the per-row prediction geometry it consumes. Because several settings depart from the COCO defaults, the figures here are internally comparable across rows but \emph{not} directly comparable to numbers published elsewhere for the same checkpoints. The OAM-TCD authors' published $0.432$ against the $0.648$ we measure for their model is the clearest instance, and reflects protocol and proposal budget (Table~\ref{tab:rpn}) rather than a different model.

\begin{table*}[!ht]
\caption{Scorer configuration. Settings marked $\dagger$ depart from the COCO defaults.\label{tab:protocol}}
\newcolumntype{K}{>{\hsize=0.55\hsize\raggedright\arraybackslash}X} \newcolumntype{V}{>{\hsize=1.45\hsize\raggedright\arraybackslash}X}
\renewcommand{\arraystretch}{1.25}
\begin{tabularx}{\textwidth}{KV}
\toprule
\textbf{Setting} & \textbf{Value} \\
\midrule
Engine & \texttt{pycocotools} \texttt{COCOeval}, unmodified; \texttt{iouType} \texttt{segm} then \texttt{bbox} \\
\cmidrule{1-2}
Unit & the $439$ whole $2048^2$ holdout tiles; AP pooled over images, not averaged per tile \\
\cmidrule{1-2}
Categories & one, \texttt{tree} (OAM-TCD \texttt{cat}~$=2$). With a single category the class mean is degenerate, which is why we write AP$_{50}$ rather than mAP$_{50}$ \\
\cmidrule{1-2}
IoU thresholds & \texttt{linspace(0.5, 0.95, 10)}, the COCO convention \\
\cmidrule{1-2}
maxDets$^{\dagger}$ & $600$, against COCO's $100$. Forced: the test tile with the most GT instances holds $450$ ($421$ crowns plus $29$ canopy groups; the densest in crowns alone holds $422$), so $100$ would censor every row. Where it binds is reported below \\
\cmidrule{1-2}
Score floor$^{\dagger}$ & $0.05$, applied to every row before scoring. COCO scores the full ranked list; this is protocol harmonisation, and it is the knob at which each method's own confidence threshold is aligned. It lowers the released default floor of Mask R-CNN (OAM-TCD) and Detectree2 from $0.2$, and since a lower floor only appends detections to the tail of the ranking, it can only improve their AP \\
\cmidrule{1-2}
Mask raster & $2048^2$ for every row, the native tile at $0.1$~m/px; masks produced at another resolution are resampled to it at the last step \\
\cmidrule{1-2}
Canopy$^{\dagger}$ & OAM-TCD's canopy class (\texttt{cat}~$=1$) enters as \texttt{iscrowd}~$=1$ GT, so a prediction landing in unlabelled canopy is ignored rather than scored. COCO's crowd rule scores intersection over detection area, so at a threshold of $0.50$ it nearly coincides with the ``more than half the prediction lies in canopy'' rule used elsewhere in this literature \\
\cmidrule{1-2}
GT & all $439$ image ids always present, so a method that predicts nothing on a tile is charged zero recall there rather than given a shrunken denominator \\
\bottomrule
\end{tabularx}
\end{table*}

\subsection{Prediction Geometry and Per-row Deviations}

No model is forced out of its own distribution; each runs in its native geometry and only the measurement is held fixed. Where we depart from a released configuration, the departure and its reason are given below. Three classes of knob are aligned across rows because the metric requires it: the confidence threshold, the per-image detection budget, and any cap that truncates the ranked list before the scorer sees it. Average precision computed over a censored ranking is not average precision.

\begin{table*}[!ht]
\begin{threeparttable}
\caption{Per-row configuration for Table~\ref{tab:oamtcd439}. \emph{Deviations} are from each method's released configuration.\label{tab:geometry}}
\newcolumntype{R}{>{\hsize=0.62\hsize\raggedright\arraybackslash}X} \newcolumntype{Y}{>{\hsize=0.68\hsize\raggedright\arraybackslash}X} \newcolumntype{D}{>{\hsize=1.70\hsize\raggedright\arraybackslash}X}
\begin{tabularx}{\textwidth}{RYD}
\toprule
\textbf{Row} & \textbf{Geometry} & \textbf{Deviations from released configuration} \\
\midrule
Mask R-CNN (OAM-TCD)\tnote{a} & whole $2048^2$ (their own test configuration) & \texttt{SCORE\_THRESH\_TEST} $0.2 \rightarrow 0.05$; \texttt{DETECTIONS\_PER\_IMAGE} $512 \rightarrow 600$; \texttt{RPN.PRE/POST\_NMS\_TOPK\_TEST} $512 \rightarrow 1000$ \\
\cmidrule{1-3}
Detectree2\tnote{b} & $1024^2$ at $0.5$ overlap, $3\times3$ subtiles, stitched & \texttt{INPUT.MIN\_SIZE\_TEST} set to its own \texttt{MIN\_SIZE\_TRAIN} of $1000$ rather than Detectron2's inherited COCO default of $800$, which would downscale our subtiles at test but not at train; stitched at the protocol floor $0.05$ (\texttt{clean\_crowns} defaults to $0.2$); as in its \texttt{stitch\_crowns}, a subtile prediction reaching within $1$~m ($10$~px) of a subtile edge is dropped before the merge, applied at edges interior to the $2048^2$ tile but not at its outer frame; \texttt{clean\_crowns} merge (IoU~$>0.7$ or containment~$>0.85$) \\
\cmidrule{1-3}
Box2Mask\tnote{c} & as the Detectree2 row & same crops, grid and stitcher, including the interior-edge rule; the architecture's cap of $100$ instances per $1024^2$ subtile is left in place \\
\cmidrule{1-3}
BoxTeacher\tnote{d} & $3\times3$ $1024^2$ subtiles at stride $512$, stitched & teacher post-NMS cap removed; canopy ignored in the classification loss \\
\cmidrule{1-3}
LACE (ours)\tnote{e} & $3\times3$ $1024^2$ subtiles at stride $512$, stitched & none (our model) \\
\bottomrule
\end{tabularx}
\begin{tablenotes}[flushleft]\footnotesize
\item[a] Released \texttt{restor/tcd-mask-rcnn-r50}, not fine-tuned; tree class only (pooling both classes: $0.634$). At the released \texttt{RPN topk} of $512$: $0.585$ (Table~\ref{tab:rpn}).
\item[b] Fine-tuned by us from detectree2's \texttt{250312\_flexi.pth} (Zenodo 15014353), using its own recipe and val-AP$_{50}$ early stopping ($6.0$ epochs).
\item[c] Box2Mask~\cite{box2mask}: Swin-L (strongest released backbone) and R-50, fine-tuned by us from COCO checkpoints with the same batch size, learning rate and selection rule as ours (val box AP$_{50}$). Output is capped at $100$ instances per $1024^2$ subtile. Swin-L hits the cap on $28.8\%$ of test subtiles (R-50: $2.7\%$), though only $1.0\%$ hold more than $100$ GT crowns. Excluding the affected tiles gains $+0.009$. R-50 stopped at epoch $6$ and selected epoch $4$; Swin-L peaked at epoch $3.5$ with three later evaluations below the peak.
\item[d] BoxTeacher~\cite{boxteacher}, Swin-B (strongest released COCO box-supervised backbone, no Swin-L has been made available), fine-tuned by us on the $900$ images (released optimiser and augmentation, validation-plateau schedule) and selected on validation box AP$_{50}$ over the same $896$ validation subtiles as Box2Mask.
\item[e] LACE with a DINO-DETR detector (Swin-L, COCO-pretrained) trained by us on the same $900$ images with boxes only, canopy ignored, and selected on validation box AP$_{50}$ over the same $896$ validation subtiles as Box2Mask; LACE masker and $\hat s$ (Section~\ref{sec:scorer}).
\end{tablenotes}
\end{threeparttable}
\end{table*}

Mask R-CNN detects only what its region proposal network (RPN) proposes, and the released configuration caps that at $512$, about half Detectron2's feature pyramid network (FPN) default, against tiles holding up to $422$ crowns. We report Detectron2's own FPN default of $1000$ (\texttt{Base-RCNN-FPN.yaml}), with the released $512$ as a sensitivity in Table~\ref{tab:rpn}. The cap starves rather than rescores: the maximum detection score is identical across settings and the additional proposals arrive entirely as low-confidence tail.

\begin{table*}[!ht]
\caption{Proposal budget of Mask R-CNN (OAM-TCD), canopy-neutral mask AP on the $439$ tiles. Recall is at IoU $0.5$; the second row is the operating point reported in Table~\ref{tab:oamtcd439}.\label{tab:rpn}}
\newcolumntype{C}{>{\centering\arraybackslash}X}
\begin{tabularx}{\textwidth}{lCCCC}
\toprule
\textbf{\texttt{RPN topk}} & \textbf{Dets/tile} & \textbf{Recall} & \textbf{AP\boldmath$_{50}$} & \textbf{AP\boldmath$_{50:95}$} \\
\midrule
$512$ (released) & $97.9$ & $0.6596$ & $0.5847$ & $0.3076$ \\
$1000$ (Detectron2 default, \textbf{reported}) & $128.6$ & $0.7441$ & $\mathbf{0.6480}$ & $\mathbf{0.3332}$ \\
\bottomrule
\end{tabularx}
\end{table*}

\subsection{Asymmetries}

The cap of $600$ detections per tile is our protocol choice. The maximum number of tree crowns in the test set is $422$, leaving comfortably enough room for accurate detections without penalty. Two models emit more than $600$ detections per tile when the cap is removed: Detectree2 emits up to $1{,}333$ at the protocol floor, and LACE up to $1{,}186$, exceeding $600$ on $212$ of the $439$ tiles. Re-scored at $1{,}200$, Detectree2 gains $0.002$ ($0.6297 \rightarrow 0.6321$) mask AP$_{50}$ and LACE $0.005$ ($0.714 \rightarrow 0.719$; AP$_{50:95}$ $0.342 \rightarrow 0.344$), which does not change any ranking order in Table~\ref{tab:oamtcd439} and shows that the residuals are a long tail of low-confidence predictions. Both are reported at the protocol cap, so the cap works against LACE rather than for it. Mask R-CNN (OAM-TCD) and Box2Mask never reach the cap, with at most $377$ and $502$ detections per tile respectively.

LACE's OAM-TCD results use a single detector training run (seed $0$). Mask R-CNN (OAM-TCD) is a single released checkpoint with no seed variation available. Detectree2, Box2Mask and BoxTeacher are trained by us but we did not run multiseed experiments. Detector seed does not enter the same-box comparisons of Section~\ref{sec:samebox}, where the boxes are held fixed and the masker fit recovers the same foreground direction at every EM seed ($\cos \geq 0.99$, Table~\ref{tab:collapse}).

DINOv3 feature extraction, LACE detector training and inference, masker fitting, and all baseline training and inference ran on an NVIDIA A100 (Modal serverless cloud\footnote{\url{https://modal.com}}); LACE's $4$~px features ($16$ offsets) and the SAM~3 zero-shot comparison ran on an Apple M4 Max graphics processing unit (GPU). Mask rendering, the learned confidence $\hat s$, the validation sweeps, subtile stitching and scoring ran on CPU, and every row of every table is scored by the same deterministic CPU code.
\end{document}